\documentclass[10pt,twocolumn,letterpaper]{article}

\usepackage{cvpr}              % To produce the CAMERA-READY version
\usepackage[dvipsnames]{xcolor}

\usepackage[accsupp]{axessibility} % Improves PDF readability for those with visual impairments.

\usepackage{graphics}
\usepackage{multicol}
\usepackage{multirow}
\usepackage{xcolor}
\usepackage{amsmath}
\usepackage{amssymb}
\usepackage{amsthm}
\usepackage{enumitem}
\usepackage{bm}
\usepackage{booktabs}
\usepackage{units}

\newtheorem{theorem}{Theorem}
\newtheorem{assumption}{Assumption}
\newtheorem{lemma}{Lemma}
\theoremstyle{definition}
\newtheorem{definition}{Definition}%[section]
\newcommand{\modelname}{\texttt{Fed$\text{U}^2$}}
\newcommand{\problem}{\texttt{FUSL}}
\newcommand{\moduleA}{\textbf{FUR}}
\newcommand{\moduleB}{\textbf{EUA}}
\newcommand{\nosection}[1]{\noindent\textbf{#1.}}
\definecolor{orange}{RGB}{255, 127, 0}

\usepackage{algorithm}
\usepackage{bbm}
\usepackage{graphicx}
\usepackage{float}
\usepackage{subcaption}
\usepackage{algorithmic}
\definecolor{cvprblue}{rgb}{0.21,0.49,0.74}
\usepackage{hyperref}
\hypersetup{pagebackref,breaklinks,colorlinks,citecolor=cvprblue}

\usepackage[capitalize]{cleveref}
\crefname{section}{Sec.}{Secs.}
\Crefname{section}{Section}{Sections}
\Crefname{table}{Table}{Tables}
\crefname{table}{Tab.}{Tabs.}

\def\paperID{2987} % *** Enter the Paper ID here
\def\confName{CVPR}
\def\confYear{2024}

\title{Rethinking the Representation in Federated Unsupervised Learning with Non-IID Data}

\author{
Xinting Liao\textsuperscript{\rm 1}, Weiming Liu\textsuperscript{\rm 1}, Chaochao Chen\textsuperscript{\rm 1}\thanks{Chaochao Chen is the corresponding author.}, Pengyang Zhou\textsuperscript{\rm 1}, Fengyuan Yu\textsuperscript{\rm 1}, Huabin Zhu\textsuperscript{\rm 1},\\
Binhui Yao\textsuperscript{\rm 1, 2}, Tao Wang\textsuperscript{\rm 2}, Xiaolin Zheng\textsuperscript{\rm 1}, Yanchao Tan\textsuperscript{\rm 3}\\
\textsuperscript{\rm 1}Zhejiang University, 
\textsuperscript{\rm 2}Midea Group, 
\textsuperscript{\rm 3}Fuzhou University\\ 
{\tt\small \{xintingliao, 21831010, zjuccc, zhoupy, fengyuanyu, zhb2000, xlzheng\}@zju.edu.cn,}\\
{\tt\small tony.yao@midea.com, tao.wang.seu@gmail.com, yctan@fzu.edu.cn}\\
}

\begin{document}
\maketitle
% \input{sec/0_abstract}    
% \input{sec/1_intro}
% \input{sec/2_formatting}
% \input{sec/3_finalcopy}
% {
%     \small
%     \bibliographystyle{ieeenat_fullname}
%     \bibliography{main}
% }
\begin{abstract}
Federated learning achieves effective performance in modeling decentralized data.
In practice, client data are not well-labeled, which makes it potential for federated unsupervised learning (\problem) with non-IID data.
However, the performance of existing \problem~methods suffers from insufficient representations, i.e., (1) representation collapse entanglement among local and global models, and (2) inconsistent representation spaces among local models.
The former indicates that representation collapse in local model will subsequently impact the global model and other local models. 
The latter means that clients model data representation with inconsistent parameters due to the deficiency of supervision signals.
In this work, we propose \modelname~which enhances generating uniform and unified representation in \problem~with non-IID data.
Specifically, \modelname~consists of flexible uniform regularizer (\moduleA) and efficient unified aggregator (\moduleB).
% and encourages uniform and unified representations.
% 
\moduleA~in each client avoids representation collapse via dispersing samples uniformly,
and \moduleB~in server promotes unified representation by constraining consistent client model updating.
To extensively validate the performance of \modelname, we conduct both cross-device and cross-silo evaluation experiments on two benchmark datasets, i.e., CIFAR10 and CIFAR100.

\end{abstract}
% WARNING: do not forget to delete the supplementary pages from your submission 
% \input{sec/X_suppl}
%-------------------------------------------------------------------------

 \section{Introduction}
%  \begin{figure}[t]
% \centering
% \includegraphics[width=\linewidth]{figs/motivation.png} 
% \caption{Motivation of \modelname.}
% \label{fig:motivation}
% \end{figure}

To meet the demands of privacy regulation,
federated learning (FL)~\cite{FedAvg} is boosting to model decentralized data in both academia and industry.
This is because FL enables the collaboration of clients with decentralized data, aiming to develop a high-performing global model without the need for data transfer.
% 
% However, most of the existing work focuses on improving performance, communication cost, and convergences in FL with non-IID data, assuming that clients contain labels to capture supervised signals. 
% % 
However, conventional FL work mostly assumes that client data is well-labeled, which is less practical in real-world applications.
In this work, we consider the problem of federated unsupervised learning (\problem) with non-IID data~\cite{fedca,fuslSurveyYangQiang}, i.e., modeling unified representation among imbalanced, unlabeled, and decentralized data.
Utilizing existing centralized unsupervised methods cannot adapt to \problem~which has non-IID data~\cite{fedu}.  
% 
% To mitigate it, there are two categories of efforts, i.e., (1) generating global supervised signals, and (2) enhancing representation generalization.
% 
To mitigate it, one of the popular categories is to train self-supervised learning models, e.g., BYOL~\cite{byol}, SimCLR~\cite{simclr}, and Simsiam~\cite{simsiam}, in clients, and 
aggregate models via accounting extremely divergent model~\cite{fedu,fedema},  knowledge distillation~\cite{fedx}, and combining with clustering~\cite{orchestra}.
However, two coupling challenges of \problem, i.e.,
\textit{\textbf{CH1}: Mitigating representation collapse entanglement}, 
and \textit{\textbf{CH2}: Obtaining unified representation spaces}, are not well considered.

The first challenge is that representation collapse~\cite{collapse_decorr} in the client subsequently exacerbates
 the representation of global and other local models. 
Motivated by regularizing Frobenius norm of representation in centralized self-supervised models~\cite{simclr_collapse,simsiam_collapse}, FedDecorr~\cite{feddecorr} tackles representation collapse with the global supervision signals in federated supervised learning.
But directly applying these methods to \problem~has three aspects of limitations.
Firstly, it relies on large data batch size~\cite{orchestra} to capture reliable distribution statistics, e.g., representation variance.
Besides, regularizing the norm of high-dimensional representations inevitably causes inactivated neurons and suppresses meaningful features~\cite{featuresupress}.
Moreover, clients cannot eliminate representation collapse entanglement by decorrelating representations for \problem~problem, once clients represent data in different representation spaces.

The second challenge refers to optimizing inconsistent client model parameters toward discrepant parameter spaces, bringing less unified representations among local models.
% (\textit{\textbf{CH2}: Obtaining inconsistent representation spaces}).
% 
Most of the existing \problem~methods aggregate participating models with the ratio of samples, i.e., FedAvg~\cite{FedAvg}.
This not only fails to tackle the client shift from global optimum to local optimum, but also brings sub-optimal results~\cite{fedprox,ldawa}.
To mitigate this, \problem~methods maintain consistency by (1) abandoning extremely divergent clients by threshold~\cite{fedema,fedu}, (2) obtaining global supervised signal via clustering client sub-clusters~\cite{orchestra,kfed}, and (3) scaling angular divergence among client models in a layer-wise way~\cite{ldawa}.
These methods either forget to adjust clients updated with inconsistent directions, or break down the performance coherence among different layers of the whole model, failing to capture unified representations.

To fill this gap, we propose a framework, i.e., \modelname, to enhance \textbf{U}niform and \textbf{U}nified representation in \problem~with non-IID data.
To tackle \textbf{CH1}, we initially devise a \textbf{flexible uniform regularizer} (\moduleA) to prevent the sample representation collapse with no regard to data distribution and client discrepancies.
In each client, \moduleA~minimizes unbalanced optimal transport divergence between client data and uniform random samples, i.e., samples from the same spherical Gaussian distribution among clients.
Thus it not only flexibly disperses local data representations toward ideal uniform distribution, but also avoids the representation collapse entanglement among clients without leaking privacy.
% 
% 
% Besides, it allows both null alignment and many-to-many alignment for optimizing towards minimal unbalanced optimal transport divergence~\cite{uotcv23,uot20}, bringing more flexible representation regularization. % 
% 
% 
% Hence every local model represents their samples in the equivalent spherical Gaussian space, mitigating the issue that the representation collapse impacts among clients.
% 
To mitigate \textbf{CH 2}, we propose \textbf{efficient unified aggregator} (\moduleB) to aggregate a global model that maintains model consistency among global optimization and different local optimizations.
Specifically, \moduleB~formulates model aggregation as a multiple-objective optimization based on the model deviation change rates of clients.
% 
% , but also dynamically manipulates the inconsistent model deviation to balanced and consistent optimal solution.
\moduleB~reduces computation by searching exact solutions in the dual formulation with alternating direction methods of multipliers.
Compared with conventional aggregation methods, we equivalently maintain consistent model updating based on client model deviation change, 
% further 
enhancing unified representations.

Summarily, we aim to enhance the representation in \problem~by mitigating representation collapse and unifying representation generalization.
(1) We enhance uniform representation by approaching data samples to spherical Gaussian distribution, which mitigates representation collapse and its subsequent entangled impacts.
(2) We enhance unified representation by constraining the consistent updating of different client models.
(3) To reach the above goals, we propose \modelname~with \moduleA~and \moduleB, which is agnostic and orthogonal for the backbone of self-supervised models.
(4) In our empirical studies, we conduct experiments on two benchmark datasets and two evaluation settings, which extensively validate the performance of \modelname.

\section{Related Work}
\subsection{Federated Unsupervised Learning}
To enhance \problem~with non-IID data~\cite{fedrane}, there are two categories of efforts, i.e.,  (1) generating global supervised signals, and (2) enhancing unified representation.
The former targets at generating global supervised signals via 
% active learning~\cite{logo}, 
local-global clustering~\cite{kfed}, and sharing data representation among clients~\cite{wu2021federated,fedca}.
But these methods either suffer from randomness in obtaining 
global supervision\cite{orchestra}, or take the risk of leaking privacy~\cite{fedu}.
% % 
The latter enhances unified representation by adapting existing unsupervised representation methods, and tackling non-IID modeling with divergence-aware model aggregation~\cite{fedu,fedema,ldawa,protofl,orchestra}.
% 
% FedCA~\cite{fedca} maintains dictionary module and alignment module to tackle representation inconsistency and misalignment, separately. 
% % 
% However, it requires additional costs for both local and global dictionaries, and leaks privacy due to aligning extra public data.
% 
% FedCA~\cite{fedca} enhances the personalization of each client by training a unique local self-supervised learning model.
% % 
% However, it relies on sharable public dataset, which inevitably will leak privacy~\cite{fedu}.
% 
Both FedU~\cite{fedu} and FedEMA~\cite{fedema} enhance the awareness of heterogeneity in federated self-supervised learning by divergence-aware predictor update rule, and adaptive global knowledge interpolation, respectively.
However, this kind of work overlooks representation collapse in non-IID clients.
%
% 
% K-Fed~\cite{kfed} captures decentralized heterogeneity by generating global centroids via local-global clustering in a one-shot communication round. 
% 
Orchestra~\cite{orchestra} utilizes local-global clustering derived from K-Fed~\cite{kfed} to guide self-supervised learning.
This brings additional cost for clustering and is fragile to random initialization.
Moreover, FedX~\cite{fedx} devises local relational loss to distill the invariance of data samples, and global relational loss to maintain client inconsistencies. 
Recently, L-DAWA~\cite{ldawa} corrects the \problem~optimization trajectory by measuring and scaling angular divergence among client models in a layer-wise way.
However,
% L-DAWA combines layer-wise parameters without the awareness of model-wise relations, i.e., the computation among layers of the same local model.
% 
it is hard to guarantee that the newly aggregated global model is still compatible and performant as a consistent model.
% 
% ProtoFL~\cite{protofl} only focuses on resolving an extreme non-IID setting, i.e., each client has one disjoint class of data, limiting its potential when the data distribution is unknown.
% 
% 
% MocoSFL~\cite{mocosfl} uses split learning to minimize the client computational burden.
% 
Differently, the proposed \modelname~enhances uniform and unified representation without the prior knowledge of unsupervised models, data distribution, and federated settings.
% 
% adapt existing self-supervised methods to \problem~by formulating a multi-objective optimization of client contribution rates, and searching global model with the maximize overall client contributions.
% 
% This enjoys the benefits of maintaining model consistency both at the model level and client level.

\subsection{Representation Collapse}
Representation collapse~\cite{simclr_collapse,simsiam_collapse} means representation vectors are highly correlated and simply span a lower-dimensional subspace, which is widely studied in metric learning~\cite{metric_collapse}, i.e., self-supervised learning~\cite{simclr_collapse}, and supervised federated learning~\cite{feddecorr}.
In federated supervised learning, FedDecorr~\cite{feddecorr} finds the dimensional collapse entanglement among server and client models, and decorrelates representations via regularizing the Frobenius norm of batch samples.
However, FedDecorr relies on large batch size and deactivates lots of neuron parameters, degrading performance once the scale of clients  increases~\cite{featuresupress}.
To avoid representation collapse entanglement in \problem, the proposed \moduleA~in \modelname~will regularize data representations to a uniform distribution that is the same among clients. 
In this way, decorrelating representation is not affected by the data sampling.
Meanwhile, the data is uniformly dispersed into the same random distribution space, 
avoiding intriguing collapse impacts of client collaboration.

\section{Method}
\begin{figure*}[t]
  \centering
  % \fbox{\rule{0pt}{0.5in} \rule{0.9\linewidth}{0pt}}
  \includegraphics[width=\linewidth]{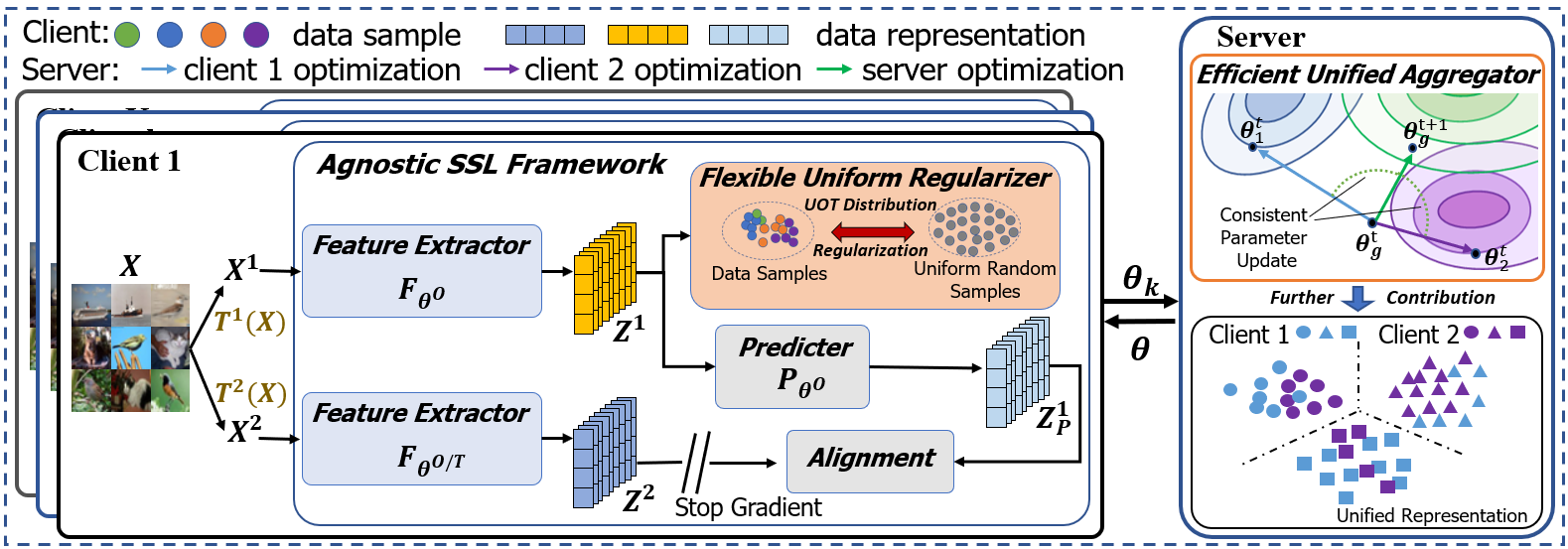}
   \caption{Framework of \modelname. For clients with agnostic self-supervised framework, \moduleA~expands non-IID data uniformly to avoid representation collapse for \problem. \moduleB~in server maintains a balanced aggregation for all client models, bringing unified representations.}
   \label{fig:model}
\end{figure*}

\subsection{Federated Unsupervised Learning Formulation}
We introduce the \problem~problem formulation and related assumptions in the following.
Empirically, we assume a dataset decentralizes among 
 $K$ clients, i.e.,  $\mathcal{D} =\cup_{k\in [K]} \mathcal{D}_k$.
Data distributions of different clients, i.e., $\mathcal{D}_k=\{ \boldsymbol{x}_{k,i}\}_{i=1}^{N_k}$, are \textit{unlabeled and non-IID} in practice.
 %
% 
% Specifically, we have $N_k$ data samples, i.e.,
% $\mathcal{D}_k=\{ \boldsymbol{x}_{k,i}\}_{i=1}^{N_k}$, for each client $k$, and the label of samples is missing and imbalanced distributed in practice.
% 
\problem~can be formulated as a global objective that seeks a collaborative aggregation among clients, i.e.,
\begin{equation}
{\operatorname{argmin}}_{\boldsymbol{\theta}} \mathcal{L}(\boldsymbol{\theta}; \boldsymbol{p})=\Sigma_{k=1}^K p_k \mathbb{E}_{\boldsymbol{x}\sim \mathcal{D}_k}[\mathcal{L}_k(\boldsymbol{\theta}; \boldsymbol{x})],
\label{eq:fusl_problem}
\end{equation}
where $\mathcal{L}_k (\cdot)$ is the unsupervised model loss at client $k$, $\boldsymbol{p}=[p_1, \dots, p_K]$, and $p_k$ represents its weight ratio. % in aggregation. 
The common aggregation approach is to assign the ratio of sample amount in client $k$ as the weight ratio, e.g., FedAvg~\cite{FedAvg}.
Nevertheless, the client with a large amount of data will dominate in aggregating, deteriorating the optimization of other clients with inconsistent local optimums~\cite{ldawa}.
Due to privacy constraints, directly aligning client local optimums with representations is forbidden~\cite{fedu,fedema}.
% , making it challenging to obtain unified and discriminative representations~\cite{fedu,fedema}.
% 
% All clients target to collaboratively model their data towards a unified representation space, via aggregating a consistent global model for the whole dataset.
% 
Therefore, 
it is necessary to account for a multi-objective optimal combination, i.e., restraining the consistency between global and local model parameters.
% to be consistent. %, which is potential for unified representations.

\subsection{\modelname~Overview}

\begin{figure}[t]
  \centering
  % \fbox{\rule{0pt}{0.5in} \rule{0.9\linewidth}{0pt}}
  \includegraphics[width=0.9\linewidth]{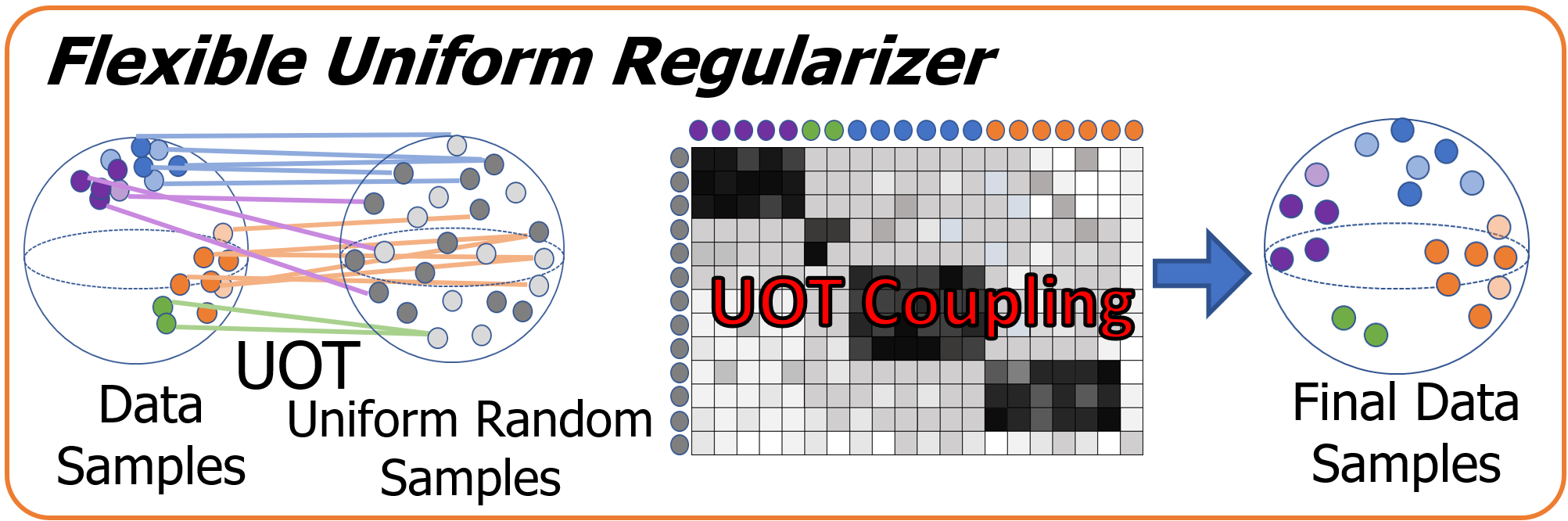}
   \caption{Example of \moduleA. Firstly, data representations collapse in part of the spherical space. Then \moduleA~flexibly maps data towards spherical Gaussian distribution with unbalanced optimal transport (UOT), dispersing data uniformly.}
   \label{fig:moduleA}
\end{figure}

To address \problem~with non-IID data, i.e., Eq.~\eqref{eq:fusl_problem}, we propose \modelname, whose framework overview is depicted in Fig.~\ref{fig:model}.
There are one server and $K$ clients in \modelname, which share the same self-supervised model, e.g., Simsiam~\cite{simsiam}, SimCLR~\cite{simclr}, and  BYOL~\cite{byol}. 
Additionally, \modelname~contains extra \textit{flexible uniform regularizer} (\moduleA) module, which mitigates representation collapse without requiring prior knowledge for \problem.
We introduce the unsupervised local modeling at each client $k$, and then illustrate the communication between clients and server.
For a batch of image data $\boldsymbol{X}$ at client $k$, we augment them with two transformations, i.e., $\boldsymbol{X}^v= T^{v}(\boldsymbol{X})$ for $T^{v}\sim \mathcal{T}$ with $\mathcal{T}$ denoting transformation set and $v\in\{1,2\}$.
% _{1}(\cdot)$ and $\mathcal{T}_{2}(\cdot)$.
% 
And feature extractor represents two views of augmented data with $d$-dimensional $l_2-$normalized representations, i.e., $\boldsymbol{Z}^{v}=\mathcal{F}_{\boldsymbol{\theta}_k}(\boldsymbol{X}^{v})$.
%
% The feature extractor $\mathcal{F}_{\boldsymbol{\theta}_k}(\cdot): \mathcal{X}\rightarrow \mathbb{R}^d$ maps a batch input data $\boldsymbol{X}_{v}$ into a $n$-dimensional vector $\boldsymbol{Z}_{v}=\mathcal{F}_{\boldsymbol{\theta}_k}(\boldsymbol{X}_{v})$ for $v\in\{1,2\}$ as feature representation.
% 
Then for each view of sample representations, we maximize its representation space via approximating uniform Gaussian distribution in \moduleA.
Meanwhile, we align two views of feature representations by predictor module (if available) and alignment module.
% 
% Compared with existing \problem~methods, client in \modelname~mitigates the representation collapse and disentangles impacts among other clients.

In one communication round, every participating client $k$ uploads its model parameters $\boldsymbol{\theta}_{k}$  to server.
Next, server with \textit{efficient unified aggregator} (\moduleB) module, first formulates the client model aggregation as a multi-objective optimization based on different client model deviation change rates, and searches for a balanced model combination.
Server further restrains client parameters in a consistent space, which not only enhances the consistency between global optimum and local optimums, but also captures unified representation for data of the same class but different clients.
This communication between server and clients iterates until the performance of \modelname~converges.

\subsection{\moduleA~for Mitigating Representation Collapse}
Representation collapse is a long-standing issue due to its intriguing phenomenon, e.g., constant collapse and partial/full dimensional collapse~\cite{simclr_collapse,byol}.
In federated learning, representation collapse not only degrades the performance of local clients, but also intricately affects the representation of global and local models~\cite{feddecorr}.
Besides, lacking labels, clients represent data samples to the space around local optimums, where decorrelating sample representations of limited client data suppresses capturing useful features~\cite{featuresupress}.
% % 
% Though FedDecorr~\cite{feddecorr} firstly analyzes and mitigates dimensional collapse for federated learning, it mainly focuses on the gradient dynamics of supervised signals.
% % 
% This cannot be directly applicable to \problem, because it decorrelates unified representation with supervision signals, in order to maintain consistent client decorrelation and prevent global dimensional collapse.
% 
% What's worse, lacking labels, clients represent data samples to the space around local optimums, where decorrelating sample representations of local limited data suppresses the effectiveness and generalization~\cite{featuresupress}.
% 他不能直接用于解决a问题,因为他需要解耦表征来增大表征的rank,但是缺乏标签的话,无法保证每一类的样本在所有client的解耦是一致的.
% 存在的问题是这样,没有对应类的信息的情况下,这个解耦学不出来,但是每个client的分布不一样,进而导致最终的
% 

Without ground truth labels, self-supervised learning not only keeps the invariance of the same sample with different augmentations, but also expands the uniformity of different representations to avoid representation collapse~\cite{ualoss}. 
Given a batch of $B$ representations, i.e., $\boldsymbol{Z}_B^1=\{\boldsymbol{z}_i^1\}_{i\in [B]}$ and $\boldsymbol{Z}_B^2=\{\boldsymbol{z}_i^2\}_{i\in [B]}$, we train the self-supervised model by minimizing the total objective as below:
\begin{equation}
\small
    \begin{gathered}
        \mathcal{L}=\mathbb{E}_{T^{1}, T^{2} \sim \mathcal{T}} 
        % \ell\left(\boldsymbol{Z}_B^{1}, \boldsymbol{Z}_B^{2}\right),
        \ell_a\left(\boldsymbol{Z}_B^{1}, \boldsymbol{Z}_B^{2}\right)+\lambda_U \left(\ell_u\left(\boldsymbol{Z}_B^{1}\right)+\ell_u\left(\boldsymbol{Z}_B^{2}\right)\right),
    \end{gathered}
    \label{eq:ssl_loss}
\end{equation}
where  $\ell_a$ and $\ell_u$ are alignment term and uniformity term, respectively.
% , i.e., $\ell\left(\boldsymbol{Z}_B^{1}, \mathbf{Z}_B^{2}\right) :=\ell_a\left(\boldsymbol{Z}_B^{1}, \boldsymbol{Z}_B^{2}\right)+\lambda_U \left(\ell_u\left(\boldsymbol{Z}_B^{1}\right)+\ell_u\left(\boldsymbol{Z}_B^{2}\right)\right)$.
% 
$\lambda_U>0$ is a hyperparameter that balances the two terms. 
The alignment term keeps data samples of the same class to be clustered, while others are separable, i.e., $   \ell_a\left(\boldsymbol{Z}_B^{1}, \boldsymbol{Z}_B^{2}\right):=\frac{1}{B} \sum_{i \in [B]}\left\|\boldsymbol{z}_i^{1}-\boldsymbol{z}_i^{2}\right\|_2^2.$
% \begin{equation}
% \small
%     \ell_a\left(\boldsymbol{Z}_B^{1}, \boldsymbol{Z}_B^{2}\right):=\frac{1}{B} \sum_{i \in [B]}\left\|\boldsymbol{z}_i^{1}-\boldsymbol{z}_i^{2}\right\|_2^2.
% \label{eq:align}
% \end{equation}
 
% 
The crucial of mitigating representation collapse is to enhance the representation uniformity~\cite{ualoss}.
% 
% However, the existing methods disperse a batch of data samples, inhabiting the utilization of the whole representation space, especially when client has insufficient data. \xenia{}
% % 
% Worse still, each client learns unlabeled data to inconsistent representation spaces, which further makes the data distribution represented in different clients incompatible.
% 
To relieve reliance on prior knowledge of client data, we regularize local sample representations to a random distribution with high entropy.
% we generalize the hypersphere uniform distribution and study a wider set of prior distributions for their effectiveness in learning representations
% 
Specifically, we select samples following the spherical Gaussian distribution, i.e., $\boldsymbol{s}\sim \mathcal{N}(0,1), s.t., \|\boldsymbol{s}\|=1$, as the prior.
In this way, mitigating representation collapse in \problem~not only avoids leaking privacy, but also disentangles the collapse impacts among clients.
% 
% Specifically, we draw a batch of random samples from the spherical Gaussian distribution, i.e., $\boldsymbol{s}\sim \mathcal{N}(0,1), s.t., \|\boldsymbol{s}\|=1$.
% 
Then \moduleA~regularizes the divergence between the data representations $\boldsymbol{Z}_B^v$ and a set of random samples following spherical Gaussian distribution ${\boldsymbol{S}_B=\{\boldsymbol{s}_i\}_{i\in [B]}}$:
\begin{equation}
\ell_u\left(\boldsymbol{Z}_B^{v}\right):= \operatorname{Div}(\boldsymbol{Z}_B^v, \boldsymbol{S}_B).
\label{eq:div}
\end{equation}
%
% Conventionally, 
% 
Thus the uniform term $\ell_u$ disperses uniformly to avoid representation collapse without repulsing in instance-based contrastive learning.
% 
% expands towards uniform prior rather than repulsing in instance-based contrastive learning, by minimizing the distribution divergence,
% 
%  we avoid the constraints of client data distributions, e.g., enhancing the variety of minority data.
% 
% Additionally, the benchmark prior constrains all clients by expanding their sample representations to the same space, further improving the generalization of global representations.
% 

% 
Since the client data is non-IID, it will break the class separation when 
strictively constraining sample representation to approach random instances~\cite{uniform_break_semantic_stuct}.
A more flexible method is to match data samples and random Gaussian samples with arbitrary or proportional masses, i.e., leaving the sampling coupling with lower uncertainties unmatched.
% samples as unmatched noisy observations.
% 
And unbalanced Optimal Transport (UOT)~\cite{uot13,uot19} is one of the effective resolutions.
UOT computes the transport mapping~\cite{aa_neurips,aa_pami,aa_www23} between two sample masses of different distributions under the soft marginal constraints, e.g., $l_2-$normalization between the predicted margin and the ground truth margin.
Given marginal constraints $\boldsymbol{a}$ and $\boldsymbol{b}$ for data and Gaussian distribution respectively, we formulate a UOT problem that searches a coupling matrix $\boldsymbol{\pi}$ with minimal distribution divergence:
\begin{equation}
\small
    \begin{aligned}
        \min _{\pi_{i, j} \geq 0} \ell_u(\boldsymbol{Z}^v) & =\operatorname{vec}(\boldsymbol{C})^{\top} \operatorname{vec}(\boldsymbol{\pi})+  \frac{\tau_a}{2}\left\|\boldsymbol{\Phi}_{\mathrm{r}} \operatorname{vec}(\boldsymbol{\pi})-\boldsymbol{a}\right\|_2^2\\
        &+\frac{\tau_b}{2}\left\|\boldsymbol{\Phi}_{\mathrm{c}} \operatorname{vec}(\boldsymbol{\pi})-\boldsymbol{b}\right\|_2^2, 
    \end{aligned}
    \label{eq:uot_raw}
\end{equation}
where cost matrix $\boldsymbol{C}_{ij} =\|\boldsymbol{Z}_{i}^v-\boldsymbol{S}_j\|^2$, and $\boldsymbol{\Phi}_{\mathrm{r}}= \boldsymbol{I}_{N} \otimes\mathbbm{1}_{N}^{\top}$ ( $\boldsymbol{\Phi}_{\mathrm{c}}= \mathbbm{1}_{M}^{\top} \otimes \boldsymbol{I}_{M}$) are indicators for row-wise (column-wise) Kronecker multiplication with $\boldsymbol{I}$ denoting identity matrix.
\nosection{Optimization} 
Denoting $\tau_a \boldsymbol{\Phi}_{\mathrm{p}}^{\top} \boldsymbol{\Phi}_{\mathrm{r}}+\tau_b \boldsymbol{\Phi}_c^{\top} \boldsymbol{\Phi}_c=\boldsymbol{Q}$ and $\operatorname{vec}(\boldsymbol{C})-\tau_a \boldsymbol{\Phi}_{\mathrm{r}}^{\top} \boldsymbol{a}-\tau_b \boldsymbol{\Phi}_c^{\top} \boldsymbol{b}=\boldsymbol{w}$, 
we rewrite Eq.~\eqref{eq:uot_raw} as a positive definite quadratic form:
\begin{equation}
\small
  \min _{\pi_{i, j} \geq 0} \ell_u(\boldsymbol{Z}^v) = \frac{1}{2} \operatorname{vec}(\boldsymbol{\pi})^\top \boldsymbol{Q} \operatorname{vec}(\boldsymbol{\pi}) + \boldsymbol{w}^{\top} \operatorname{vec}(\boldsymbol{\pi})+\boldsymbol{\Omega},
\end{equation}
where constant $\boldsymbol{\Omega}=\frac{1}{2} (\tau_a \boldsymbol{a}^{\top}\boldsymbol{a}+ \tau_b \boldsymbol{b}^{\top}\boldsymbol{b})$.
Next we optimize $\boldsymbol{\pi}$ via steepest gradient descent as bellow:
\begin{equation}
\small
\operatorname{vec}\left(\boldsymbol{\pi}^{(\text {new })}\right)=\max \left(0, \operatorname{vec}\left(\boldsymbol{\pi}^{\text {(old })}\right)-\eta^* \frac{\partial \ell_u(\boldsymbol{Z}^v)}{\partial \operatorname{vec}\left(\boldsymbol{\pi}^{(\text {old })}\right)}\right),
\end{equation}
with $\eta^*=\frac{\left(\boldsymbol{Q}_{\operatorname{vec}}\left(\boldsymbol{\pi}^{(\text {old })}\right)+\boldsymbol{w}\right)^{\top}\left(\boldsymbol{Q}_{\operatorname{vec}}\left(\boldsymbol{\pi}^{(\text {old })}\right)+\boldsymbol{w}\right)}{\left(\boldsymbol{Q}_{\operatorname{vec}}\left(\boldsymbol{\pi}^{(\text {old })}\right)+\boldsymbol{w}\right)^{\top} \boldsymbol{Q}\left(\boldsymbol{Q} \operatorname{vec}\left(\boldsymbol{\pi}^{(\text {old })}\right)+\boldsymbol{w}\right)}$.
Finally, we obtain the uniform UOT divergence by taking the optimal $\boldsymbol{\pi}^{*}$ back to Eq.~\eqref{eq:uot_raw}.
\moduleA~minimizes UOT divergence to
regularize data samples approaching the spherical Gaussian distribution.
Note that the spherical Gaussian distribution maximizes its entropy and distributes its samples uniformly.
The mapped data representations enjoy the above nice properties of spherical Gaussian distribution and further mitigate the representation collapse entanglement.

\subsection{\moduleB~for Generalizing Unified Representation}
Due to non-IID client data, clients optimize to their local optimums with inconsistent model parameters, causing inconsistent even conflicting model deviations from server to clients.
Without the guidance of supervision signals, i.e., data labels, this problem further exacerbates in representing data of the same class but different clients, towards inconsistent spaces.
% 
% % 
Thus it is vital to constrain the consistency among client models in parameter spaces, which further guarantees unified representations.
% 
% L-DAWA~\cite{ldawa} computes angular divergence from global model to local model as the layer-wise aggregating ratio.
% % % 
% This scales the client contributions, and achieves smoother control of the trajectory of model optimization.
% % 
% However, the layer-wise aggregated model not only fails to guarantee a compatible and performant model, but also overlooks the case that local clients optimize to conflicting directions.
% 

% The model deviations from server to different clients, i.e., $u_k(\boldsymbol{\theta}^{t}_g)=\| \boldsymbol{\theta}^{t}_g - \boldsymbol{\theta}^{t}_k\|^2$, are inconsistent, 

In round $t$, the impact of global aggregation on $k$-th local optimization can be measured with the model deviation change rate, i.e., 
\begin{equation}
\small
\begin{aligned}
c_k \left(\eta, \boldsymbol{d}^t\right)=\frac{u_k\left(\boldsymbol{\theta}^{t}_g\right)-u_k\left(\boldsymbol{\theta}^{t+1}_g\right)}{{u}_k \left(\boldsymbol{\theta}^{t}_g\right)}
\approx \eta_g \nabla \log {u}_k(\boldsymbol{\theta}^t_g) \boldsymbol{d}^t,
\end{aligned}
\end{equation}
where $u_k(\boldsymbol{\theta}^{t}_g)=\| \boldsymbol{\theta}^{t}_g - \boldsymbol{\theta}^{t}_k\|^2$ is the model deviation from server to client $k$~\cite{scaffold,fedmgda+}, and global model optimization is $\boldsymbol{\theta}_g^{t+1}=\boldsymbol{\theta}_g^{t} - \eta \boldsymbol{d}^t$ with updating direction $\boldsymbol{d}^t$ and step size $\eta$.
Overlooking inconsistent model deviations, global aggregated model inevitably gets close to a subset of clients while deviating from others.
It corresponds that 
clients get close to the global model increase the model deviation change rate, and clients away decrease it~\cite{fedmgda+,fedmdfg}.
Motivated by this, we seek the clients with the worst model deviation change rate, and correct the global optimization with a direction maximizing the overall worst model deviation change rate.
% % 
This can be formulated as a multi-objective optimization, which benefits for mitigating the inconsistencies and conflicts among clients~\cite{fedmgda+,mgda}, i.e., 
\begin{equation}
\label{eq:mtl_cc}
\small
\begin{aligned}
&\max _{\boldsymbol{d}^t} \min _{\boldsymbol{p} \in \mathbb{S}^K} {\eta}_g \sum_{k=1}^{K} p_k \nabla \log  {u}_k(\boldsymbol{\theta}^t_g) \boldsymbol{d}^t,\\
&s.t. \|\boldsymbol{d}^t\|^2 \leq 1, \boldsymbol{p}^\top \boldsymbol{1} =1, p_k \ge 0, 
\end{aligned}
\end{equation}
% and $\mathbb{S}^K=\{\boldsymbol{p}\in \mathbb{R}^K | \boldsymbol{p}^\top \boldsymbol{1}=1\}$.
% 
where $\bm{p}$ denotes the weights for different clients.
%.
% Then we introduce the optimization details on Eq.\eqref{eq:mtl_cc} below.

\nosection{Optimization}
Adding the constraints as Lagrange multipliers, Eq.~\eqref{eq:mtl_cc} can be rewritten as:
\begin{equation}
\small
\max _{\boldsymbol{d}^t} \min _{\boldsymbol{p}} J=\eta_g \left< \sum_{k=1}^{K} p_k \nabla \log  {u}_k(\boldsymbol{\theta}^t_g), \boldsymbol{d}^t\right>  - \frac{\phi}{2}(\|\boldsymbol{d}^t\|^2 -1).
\label{eq:mtl_cc_La}
\end{equation}
Differentiating Eq.~\eqref{eq:mtl_cc_La} with regarding to $\boldsymbol{d}^t$, we have $\boldsymbol{d}^*= \frac{\eta_g}{\phi} {(\nabla\log  {\boldsymbol{u}})}^\top \boldsymbol{p}$, where $ {\boldsymbol{u}}= ( {u}_1(\boldsymbol{\theta}_g^t),\dots, {u}_K (\boldsymbol{\theta}_g^t))$.
Taking back to Eq.~\eqref{eq:mtl_cc_La}, we can obtain its strong dual form with dual variable $\boldsymbol{p}$:
\begin{equation}
\small
\boldsymbol{J} =\min _{\boldsymbol{p}} \frac{\eta^2_g}{2\phi} \|{(\nabla\log  {\boldsymbol{u}})}^\top \boldsymbol{p}\|^2 
=\min _{\boldsymbol{p}} \frac{\eta^2_g}{2\phi} \boldsymbol{p}^\top \boldsymbol{G} \boldsymbol{p},
\label{eq:mtl_cc_dual}
\end{equation}
% \begin{equation}
% \begin{aligned}
% \boldsymbol{J} &=\min _{\boldsymbol{p}} \frac{\eta^2}{2\phi} \|{(\nabla\log  {\boldsymbol{u}}(\boldsymbol{\theta}^t))}^\top \boldsymbol{p}\|^2 \\
%  &=\min _{\boldsymbol{p}} \frac{\eta^2}{2\phi} \boldsymbol{p}^\top \boldsymbol{G} \boldsymbol{p},\\
% \end{aligned}
% \label{eq:mtl_cc_dual}
% \end{equation}
where $\boldsymbol{G}=(\nabla\log  {\boldsymbol{u}})^\top (\nabla\log  {\boldsymbol{u}})$.
Then we can rewrite it as an augmented Lagrangian form,
\begin{equation}
\small
\begin{aligned}
\boldsymbol{J} =\min _{\boldsymbol{p}} \frac{\eta^2_g}{2\phi} \boldsymbol{p}^\top \boldsymbol{G} \boldsymbol{p}+\mu (\boldsymbol{p}^\top \boldsymbol{1}-1) +\frac{\rho}{2}\|\boldsymbol{p}^\top \boldsymbol{1}-1\|^2, \\
\end{aligned}
\label{eq:famo_dual_aug}
\end{equation}
where $\mu$ denotes the Lagrange multipliers.
This can be iteratively solved by alternating direction method of multipliers (ADMM) algorithm~\cite{admm,aa_admm}, i.e., fixing $\mu$ to optimize $\boldsymbol{p}$, and vice versa:
% \begin{equation}
% \begin{aligned}
% \left\{\begin{array}{ll}
%      & \frac{\partial{J}}{\partial{\boldsymbol{p}}} =\frac{\eta^2}{\phi}\boldsymbol{G}\boldsymbol{p} + \mu\boldsymbol{p}  + \rho (\boldsymbol{p}-1)=0 \\
%      &  \frac{\partial{J}}{\partial{\mu}}=\boldsymbol{p}^\top \boldsymbol{1}-1=0
% \end{array}\right.
% \end{aligned}
% \end{equation}
% 
\begin{equation}
\small
\begin{aligned}
    \left\{\begin{array}{ll}
     &\boldsymbol{p} = \max (0,(\frac{\eta^2_g}{\phi}\boldsymbol{G}+\rho\boldsymbol{I})^{-1} (\rho\boldsymbol{I} -\mu\boldsymbol{I}) )\\
     & \mu \leftarrow \mu+ \rho (\boldsymbol{p}^\top \boldsymbol{1}-1)
\end{array}\right.
\end{aligned}
\end{equation}
The ADMM iteration guarantees exact solution in minimal computation complexity, then the global model updates towards $\boldsymbol{d}^*$ with step size $\eta$.

\begin{theorem}[Optimization consistency of model deviations]
Rethinking the Lagrangian of dual form in Eq.~\eqref{eq:mtl_cc_dual}, 
\begin{equation}
\boldsymbol{J} =\min _{\boldsymbol{p}} \frac{\eta^2_g}{2\phi} \|{(\nabla\log  {\boldsymbol{u}})}^\top \boldsymbol{p}\|^2 + \lambda \boldsymbol{p}^\top \boldsymbol{1},
\label{eq:damo_dual_add}
\end{equation}
it holds $\nabla \log( {u}_i\left(\boldsymbol{\theta}^{t}_g\right)) =\nabla \log( {u}_j\left(\boldsymbol{\theta}^{t}_g\right))$, $\forall i \neq j\in [K]$.
\begin{proof}
We provide the proof details in Appendix A.1.
\end{proof}
\end{theorem}

After the convergence of global and local optimization, \moduleB~balances the model deviation change rate among all clients, 
making the global aggregation improves all model equivalently.
Therefore, all models optimize towards a consistent parameter spaces, obtaining unified representation.

\subsection{Overall Algorithm and Convergence Analysis}
We describe the overall algorithm of \modelname~in Algo.~\ref{alg:fl_model}.
% 
% Steps 1:10 are the main procedure of \modelname.
% 
In detail, the server collaborates with clients in steps 1:10.
After collecting participating client models in step 8, server uses \moduleB~to reach a consistent model updating and obtain unified representations.
The client executes self-supervised modeling in steps 11:21, where \moduleA~enhances uniform representations to avoid collapse entanglement in step 17.

\nosection{Convergence Analysis}
In the following, we
 take four mild assumptions~\cite{zhihua_convergence}, and
provide the generalization bounds of model divergence and overall convergence error.

\begin{assumption}
\label{asu:l-lip}
Let $F_k(\boldsymbol{\theta}_k)$ be the expected model objective for client $k$, and assume
    $F_1, \cdots, F_K$ are all L-smooth, i.e., for all $\boldsymbol{\theta}_k$, $F_k(\boldsymbol{\theta}_k) \leq F_k(\boldsymbol{\theta}_k)+(\boldsymbol{\theta}_k- \boldsymbol{\theta}_k)^\top \nabla F_k(\boldsymbol{\theta}_k)+\frac{L}{2}\|\boldsymbol{\theta}_k- \boldsymbol{\theta}_k\|_2^2$.
\end{assumption}
\begin{assumption}
\label{asu:cvx}
   Let $F_1, \cdots, F_N$ are all $\mu$-strongly convex: for all $\boldsymbol{\theta}_k$, $F_k(\boldsymbol{\theta}_k) \geq F_k(\boldsymbol{\theta}_k)+(\boldsymbol{\theta}_k- \boldsymbol{\theta}_k)^\top \nabla F_k(\boldsymbol{\theta}_k)+\frac{\mu}{2}\|\boldsymbol{\theta}_k- \boldsymbol{\theta}_k\|_2^2$.
\end{assumption}

\begin{assumption}
\label{asu:bound_var}
Let $\xi^t_k$ be sampled from the $k$-th client's local data uniformly at random. The variance of stochastic gradients in each client is bounded: $\mathbb{E}\left\|\nabla F_k\left(\boldsymbol{\theta}_k^{t}, \xi^t_k\right)-\nabla F_k\left(\boldsymbol{\theta}_k^{t}\right)\right\|^2 \leq \sigma_k^2$. % $(\forall k=1, \cdots, K)$.
\end{assumption}

\begin{assumption}
\label{asu:bound_grad_norm}
The expected squared norm of stochastic gradients is uniformly bounded, i.e., $\mathbb{E}\left\|\nabla F_k\left(\boldsymbol{\theta}_k^{t}, \xi^t_k\right)\right\|^2 \leq V^2$ for all $k=1, \cdots, N$ and $t=1, \cdots, T-1$
\end{assumption}

\begin{lemma}[Bound of Client Model Divergence]
\label{lemma:div}
With assumption~\ref{asu:bound_grad_norm}, $\eta_t$ is non-increasing and $\eta_t< 2\eta_{t+E}$ (learning rate of t-th round and E-th epoch) for all $t\geq 0$, there exists $t_0 \leq t$, such that $t-t_0\leq E-1$ and $\boldsymbol{\theta}^{t_0}_k= \boldsymbol{\theta}^{t_0}$ for all $k\in [N]$. It follows that 
\begin{equation}
\label{supp_eq:var_bound}
\mathbb{E} \left[ \sum_k^K p_k \|\boldsymbol{\theta}^{t}- \boldsymbol{\theta}^{t}_k\|^2 \right]\leq 4\eta_t^2 {(E-1)}^2 V^2.
\end{equation}
% 
% \begin{lemma}[Bound of Client Model Divergence]
% \label{lemma:div}
% With assumption~\ref{asu:bound_grad_norm}, $\eta_t$ is non-increasing and $\eta_t< 2\eta_{t+E}$ for all $t\geq 0$, there exists $t_0 \leq t$, such that $t-t_0\leq E-1$ and $\boldsymbol{\theta}^{t_0}_k= \boldsymbol{\theta}^{t_0}$ for all $k\in [N]$. It follows that 
% \begin{equation}
% \label{eq:var_bound}
% \mathbb{E} \left[ \sum_k^K p_k \|\boldsymbol{\theta}^{t}- \boldsymbol{\theta}^{t}_k\|^2 \right]\leq 4\eta_t^2 {(E-1)}^2 V^2.
% \end{equation}
\begin{proof}
We provide the proof details in Appendix A.2.
\end{proof}
\end{lemma}

\begin{theorem}[Convergence Error Bound]
\label{th:convergence}
Let assumptions 1-4 hold, and $L, \mu, \sigma_k, V$ be defined therein.
Let $\kappa=\frac{L}{\mu}, \gamma=\max\{8\kappa, E\}$ and the learning rate $\eta_t=\frac{2}{\mu(\gamma+t)}$.
The \modelname~with full client participation satisfies 
$$
\mathbb{E}\left[F\left({\overline{\boldsymbol{\theta}}}^t\right)\right]-F^* \leq \frac{\kappa}{\gamma+t}\left(\frac{2 B}{\mu}+\frac{\mu(\gamma+1)}{2} \|\boldsymbol{\theta}^t - \boldsymbol{\theta}^{*}\|^2\right),
$$
where $B=4(E-1)^2V^2 +K+2\Gamma$.
\begin{proof}
We provide the proof details in Appendix A.3.
\end{proof}
\end{theorem}

\begin{algorithm}[tb]
    \caption{Training procedure of \modelname}
    \label{alg:fl_model}
    \textbf{Input}: Batch size $B$, communication rounds $T$, number of clients $K$, local steps $E$, dataset $\mathcal{D} =\cup_{k\in [K]} \mathcal{D}_k$\\
    \textbf{Output}: Global model $\boldsymbol{\theta}^T$
    \begin{algorithmic}[1]
        \STATE \textbf{Server executes():}
        \STATE Initialize 
 $\boldsymbol{\theta}^{0}$ with random distribution 
        % \STATE Send model parameters to all participating clients
        \FOR{$t=0,1,...,T-1$}
            \FOR{$k=1,2,...,K$ \textbf{in parallel}} 
                \STATE Send $\boldsymbol{\theta}^t$ to client $k$ 
                \STATE $\boldsymbol{\theta}_k^{t+1} \leftarrow$ \textbf{Client executes}($k$, $\boldsymbol{\theta}^t$)
            \ENDFOR
        \STATE \moduleB~optimize Eq.~\eqref{eq:mtl_cc_dual} for $\boldsymbol{p}^*$ and   update global model $\boldsymbol{\theta}^{t+1}$ with optimal direction  $\boldsymbol{d}^{t}$ in Eq.~\eqref{eq:mtl_cc_La} 
        \ENDFOR
        \STATE \textbf{return} $\boldsymbol{\theta}^T$
        % \STATE
        \STATE  \textbf{Client executes}($k$, $\boldsymbol{\theta}^t$)\textbf{:}
        \STATE Assign global model to the local model $\boldsymbol{\theta}_k^t \leftarrow \boldsymbol{\theta}^t$
        \FOR{each local epoch $e= 1, 2,..., E$}
            \FOR{batch of samples $\boldsymbol{X}_{k,B} \in \mathcal{D}_{k}$}
            \STATE Augment samples $\boldsymbol{X}_{k,B}^v = T^{v}_{v\in \{1,2\}}(\boldsymbol{X}_{k,B})$ 
                \STATE Feature extraction $\boldsymbol{Z}_{k, B}^{v} \leftarrow \mathcal{F}_{\boldsymbol{\theta}_k^e} (\boldsymbol{X}_{k, B}^v)$ 
                \STATE \moduleA~enhances the uniformity of $\boldsymbol{Z}_{k, B}^v$ by Eq.~\eqref{eq:uot_raw}
                \STATE Compute total loss in Eq.~\eqref{eq:ssl_loss} and update $\boldsymbol{\theta}_k^e$ 
            \ENDFOR
        \ENDFOR
        \STATE \textbf{return} $\boldsymbol{\theta}_k^E$ to server
    \end{algorithmic}
\end{algorithm}

\section{Experiments}
\subsection{Experimental Setups}
\nosection{Datasets}We adopt two benchmark datasets, i.e., CIFAR10 and CIFAR100~\cite{cifardata}, to evaluate \modelname.
Both datasets have 50,000 training samples and 10,000 test samples, but differ in the number of classes. 
% 
% TinyImageNet contains 200 classes, with 100,000 training samples and 10,000 testing samples.
% 
Following FedEMA~\cite{fedema}, we simulate non-IID data distribution in $K$ clients
% using label heterogeneity.
% % 
% Specifically, we evaluate the performance of \modelname~
by assuming class priors follow the Dirichlet distribution parameterized with non-IID degree, i.e., $\alpha$~\cite{dirichlet_sample}.  
The smaller $\alpha$ simulates the more non-IID federated setting.
% % 
We conduct extensive experiments in both cross-silo ($K=10$) and cross-device ($K=100$) settings to validate performance generalization.

\nosection{Comparison Methods}
% ref Federated Learning with Label Distribution Skew via Logits Calibration
% 
We compare \modelname~with three categories of approaches 
, i.e., (1) combining the existing centralized self-supervised model with FedAvg~\cite{FedAvg}: \textbf{FedSimsiam}, \textbf{FedSimCLR}, and \textbf{FedBYOL}, (2) the state-of-the-art \problem~methods:  \textbf{FedU}~\cite{fedu}, \textbf{FedEMA}~\cite{fedema}, \textbf{FedX}~\cite{fedx}, \textbf{Orchestra}~\cite{orchestra}, and \textbf{L-DAWA}~\cite{ldawa}, and (3) adapting existing federated supervised learning models solving representation collapse to \problem: \textbf{FedDecorr}~\cite{feddecorr}. 
Firstly, we evaluate the representation performance of the above methods with their best-performing models on both cross-silo and cross-device settings on CIFAR10 and CIFAR100 ($\alpha=0.1$).
Secondly, we study the effectiveness of different methods with the same model, i.e., BYOL, for different non-IID degrees~\cite{hyperfed}, i.e., $\alpha=\{0.1, 0.5, 5\}$.
% 
% The first evaluation aims to compare the representation performance of  FedAvg~\cite{FedAvg} with conventional centralized self-supervised learning, i.e., FedSimsiam, FedBYOL, FedSimCLR, and the state-of-the-art \problem~methods with their best-performing models, i.e., 
% representation 
% i.e., (1) the pre-trained model for the substream task in the most complicated task (i.e., $\alpha=0.1$), and (2) 
% the performance generalization for different heterogeneities.
% % 
% For the former, we compare the best performant model of existing \problem~methods, i.e., FedAvg with conventional centralized self-supervised learning, e.g., FedSimsiam, FedBYOL, FedSimCLR, the state-of-the-art \problem~method, e.g., FedU~\cite{fedu}, FedEMA~\cite{fedema}, FedX~\cite{fedx}, Orchestra~\cite{orchestra}, and L-DAWA~\cite{ldawa},
% and the mitigating representation collapse method, FedDecorr in \problem.
% % 
% For the latter, we choose the same self-supervised model, i.e., BYOL, for all \problem~methods, in order to fairly study performance generalization among different non-IID settings.
% 
% We provide the above method details in Appendix \xenia{todo}. 
% 
We evaluate the pre-trained encoder model via the accuracy of KNN~\cite{knn}, standard linear probing~\cite{linear_prob}, and semi-supervised methods (i.e., fine-tuning 1\% and 10\% labeled data).
% '
All of the above metrics illustrate better performance when the values are higher.

\nosection{Implemental Details}
We conduct image augmentation for SimCLR, BYOL, and SimSiam, following their original papers.
% 
% For the existing baseline methods, we follow the transformation used in their original papers.
% ref Adapt to Adaptation: Learning Personalization for Cross-Silo Federated Learning
% ref Personalized Federated Learning with Contextualized Generalization
% 
% We conduct all of the experiments on one NVIDIA RTX 3090 GPU with 24Gb Memory.
% 
% 
We adopt ResNet18~\cite{he2016deep} as an encoder module, choose Projector/Predictor architectures like original papers,
and optimize each model 5 local epochs per communication round until converging.
% % 
% % 
% We 
We set all datasets with batch size as 128 and embedding dimension as 512.
To obtain fair comparisons, we conduct every experiment for each method with its best hyper-parameters, and report the average result of 3 repetitions.
We choose Adam~\cite{adam} as the optimizer for each local model, and SGD~\cite{sgd} for updating the global model.
We set the uniformity effect $\lambda_U=0.1$, the soft margin constraints $\tau_a= \tau_b=0.8$, the coefficient of constraints $\phi=0.1$, and the coefficient in ADMM $\rho=1$.
% in Eq.~\eqref{eq:ssl_loss} in $\{0, 0.01, 0.1, 0.2, 0.5, 1\}$.
% \xenia{tune hyper-param for uot $\eta$ in global aggregation}

\begin{table*}[t]
\caption{Accuracy (\%) of linear probing (LP), fine-tuning (FT) 1\%, and 10\% labeled data on CIFAR10 and CIFAR100 ($\alpha=0.1$).}
\resizebox{0.98\linewidth}{!}{
\begin{tabular}{lllllllllllll}
\toprule
Dataset    & \multicolumn{6}{c}{CIFAR10}                                                                                                                                                                           & \multicolumn{6}{c}{CIFAR100}                                                                                                                                                                           \\
\hline
Setting $\alpha=0.1$  & \multicolumn{3}{c}{Cross-Device (K=100)}                                                                               & \multicolumn{3}{c}{Cross-Silo (K=10)}                                    & \multicolumn{3}{c}{Cross-Device (K=100)}                               & \multicolumn{3}{c}{Cross-Silo (K=10)}                                                                              \\
\hline
  Method \textbackslash Evaluation        & \multicolumn{1}{c}{LP}    & \multicolumn{1}{c}{FT 1\%} & \multicolumn{1}{c}{FT 10\%} & \multicolumn{1}{c}{LP}    & \multicolumn{1}{c}{FT 1\%} & \multicolumn{1}{c}{FT 10\%} & \multicolumn{1}{c}{LP}    & \multicolumn{1}{c}{FT 1\%} & \multicolumn{1}{c}{FT 10\%} & \multicolumn{1}{c}{LP}    & \multicolumn{1}{c}{FT 1\%} & \multicolumn{1}{c}{FT 10\%} \\
\hline
FedSimsiam &     60.49                     &  44.45             &     70.46                              &     70.61 & 57.60 & 69.88                              &      31.91                     &    12.58                               &           37.33                        &     49.81 & 21.64 & 43.08                            \\
FedDecorr-Simsiam  & 43.18 & 35.15  & 58.68 & 74.56 & 65.21  & 80.10  & 17.09 & 5.36&  20.60  & 47.93  & 20.53 & 45.21\\
\modelname-Simsiam                         &   \textbf{68.50}  & \textbf{56.43}  & \textbf{75.33} 
&       \textbf{84.92}	 & \textbf{77.11} & 	\textbf{85.21}&
                                                    \textbf{35.59} & \textbf{13.08} & \textbf{38.22}                                                         &   \textbf{56.55}  & \textbf{31.42}  & \textbf{48.75}  \\
                                                    \hline
FedSimCLR  &         65.76     &     51.18                         &                      	68.33       &      75.65 & 62.86 & 76.15      &                 37.09                 &   11.73                          &               	31.97               &    51.62 & 19.47 & 41.60                             \\
L-DAWA-SimCLR     &       65.63	           &                       49.66            &                         	69.89                &          75.48           & 	63.4                                     &         	78.66               &           41.28 &                      	13.52                          &   	36.56                                  &                 51.11	  &                     21.07	                    &  45.02                                   \\
FedX-SimCLR       &   67.33	            &             49.96	                     &  70.18                                      &        78.29                   &    65.03	                              &       79.43                             &        38.11    &                         	11.18                     &            	33.96                           &                 51.67          &      19.65	                         &                      42.38                 \\
    
\modelname-SimCLR          &    66.49  & \textbf{51.77}  & \textbf{70.76}      & \textbf{82.37}                                                                                                   &    \textbf{69.84}                    & \textbf{82.39}   & \textbf{41.56}         &   	\textbf{14.32}	                 &                      \textbf{36.90}                     &   \textbf{56.56}	 &                           \textbf{26.11}	        &      \textbf{47.99}                                                           \\
\hline
FedBYOL    &        61.46	  &    54.36 &   	74.01                                                 &      83.29 & 74.04 & 81.40                                    &         28.27	                  &               	10.43&   	34.90  &               48.78 & 19.79 & 42.82              \\
FedU-BYOL       &       60.15                    &          53.53	              &              74.62                      &     82.33             &                      	69.24                     &       83.37                         &          28.09                &        	10.46              &                     36.06                 &           58.02            &                       28.38         &       	48.12                           \\      
FedEMA-BYOL     &       62.27     &                54.91                                                   &       	74.76                &    82.17                              &  71.37 	                          &        83.78                   &      28.40 	            &             10.63	          &    35.62                   &            57.25                       &      	30.03          &     	50.33                                                 \\

Orchestra  &      38.66	     &          41.62	               &                   62.97                &       83.53               &  78.44                                 &          	85.40                  &       17.91  & 6.96  & 23.40  &          51.31                       & 26.36                   &         48.85                                                              \\
\modelname-BYOL          &        \textbf{67.62} & \textbf{54.74} & \textbf{74.93}         
&         \textbf{85.58} & 	\textbf{78.64} & 	\textbf{86.24}            
&   \textbf{38.09} & \textbf{13.16} & \textbf{36.87}                 
&  \textbf{59.71}  & \textbf{34.83} & \textbf{53.87}  \\
\bottomrule
\end{tabular}
}
\label{tb:pretrained}
\end{table*}

\begin{table}[t]
\caption{KNN accuracy (\%) of different $\alpha$ on CIFAR10 and CIFAR100 for cross-silo settings.}
\resizebox{\columnwidth}{!}{
\begin{tabular}{lcccccccccccc}
\toprule
Dataset    & \multicolumn{3}{c}{CIFAR10}                 & \multicolumn{3}{c}{CIFAR100}                \\
\hline
Method \textbackslash $\alpha$      & 0.1    & 0.5   & 5   & 0.1    & 0.5   & 5    \\
\hline
% FedSimsiam & \multicolumn{1}{c}{88.88} & \multicolumn{1}{c}{88.88}         & \multicolumn{1}{c}{88.88}         & \multicolumn{1}{c}{88.88} & \multicolumn{1}{c}{88.88}         & \multicolumn{1}{c}{88.88}      \\
FedBYOL      &  76.12 &   77.23     & 82.71     &  38.13  & 	43.93   &   45.30        \\
FedU-BYOL       &   79.09& 	79.68   &    	82.75             &           51.31&  	51.81&	52.05     \\
% FedSimCLR  &        &       &     &        &       &        \\
FedEMA-BYOL     &  80.32 & 82.01   & 82.80  & 53.18  & 53.18  & 53.28       \\
FedDecorr-BYOL     &    76.76& 	79.66  &  	81.09  &  49.87&   49.54   &   	52.07        \\
L-DAWA-BYOL     &  65.40  & 66.17& 82.82 & 23.09  & 51.10 & 52.25      \\
FedX-BYOL      & 50.94      &  	40.96&	41.05        &  15.83  &    	16.35  &    	16.89  \\
Orchestra  &    79.25 &  76.78 & 	76.30    &    38.52   &  	37.17     &  34.93        \\
\hline
\modelname-\moduleA-BYOL   &    81.04    &   82.18    &  83.45  &  53.45      &      53.86 &     54.34  \\ 
\modelname-\moduleB-BYOL       &     80.93
  & 82.14      &  84.01   &     53.20   & 53.72      &     54.61  \\
\modelname-BYOL      &   \textbf{81.39}       &  \textbf{82.21}     &   \textbf{84.79}  &  \textbf{53.87}      &  \textbf{54.07}     &   \textbf{55.06}    \\
\bottomrule
\end{tabular}
}
\label{tb:noniid_generalization}
\end{table}

\subsection{Experimental Results}

\nosection{Representation Performance Comparison}
Firstly, we follow the existing \problem~methods~\cite{fedema,orchestra,ldawa}, to evaluate the performance of pre-trained models learned in Tab.~\ref{tb:pretrained}.
We group the state-of-the-art methods in terms of their best-performing models, i.e., Simsiam, SimCLR, and BYOL.
\textbf{For the first group}, we can observe that FedDecorr performs the worst, especially on CIFAR100 cross-device task.
It indicates that directly avoiding collapse via decorrelating a batch of data representations is unsuitable for \problem~with limited data and severely heterogeneous data distribution.
\textbf{In terms of the second group}, compared with FedX, L-DAWA performs better on CIFAR100,  while is less competitive on CIFAR10.
We can conclude that: (1) L-DAWA can better control model divergence when clients have inconsistent optimums, and (2) L-DAWA fails to obtain discriminative representations since it takes no action to representation collapse.
\textbf{On mentioned the third group}, Orchestra captures global supervision signals to guide data representation, whose effectiveness suffers from randomness.
\textbf{In general}, directly combining existing self-supervised model with FedAvg cannot tackle \problem~with non-IID data well.
Cross-device simulation on CIFAR100 is so challenging that some existing methods fail dramatically.
Moreover, \modelname~is agnostic to self-supervised model and performs better than existing work, which validates the superiority of enhancing uniform and unified representations.

\nosection{Effect of Heterogeneity on Generalization}
Next, we report the KNN-accuracy of cross-silo methods on CIAFR10 and CIFAR100 in Tab.~\ref{tb:noniid_generalization}, for validating the performance generalization.
We choose the same model, i.e., BYOL, to be comparable among all \problem~methods.
We can discover that: (1) Most \problem~methods increase their performance when the non-IID degree $\alpha$ increases, and the performance variances among different methods increase with the decreasing of $\alpha$.
(2) FedEMA-BYOL is not sensitive to the non-IID degrees, while Orchestra behaves on opposite. 
This states that capturing global supervision signals to guide local representation suffers from clustering randomness.
(3) \modelname~performs the best among all tasks, even in $\alpha=0.1$, illustrating its performance generalization.

\nosection{Ablation Studies}
In Tab.~\ref{tb:noniid_generalization}, we also consider two variants of \modelname: 
% (1). \modelname~using fixed and shared prototypes with contrastive learning in Euclidean space (Cosine Distance), i.e., \modelname-$\mathbb{E}$,
(1) \modelname~removes \moduleA, i.e., \modelname-\moduleA,
(2) \modelname~removes \moduleB, i.e., \modelname-\moduleB, to study the effect of each module.
From Tab.~\ref{tb:noniid_generalization}, we can see that either applying \moduleA~or \moduleB~can enhance representations, since they have better performance than the existing \problem~methods.
Compared with \modelname, \modelname-\moduleA~and \modelname-\moduleB~drop KNN accuracy slightly, validating the effectiveness of tackle two challenges, i.e., representation collapse entanglement, and generating unified representations.
\modelname-\moduleA~performs better than \modelname-\moduleB~when $\alpha=0.1$, while gets worse when $\alpha=5$.
% 
% This means that \moduleB~is more effective in serve non-IID data, but \moduleA~works better less non-IID data.

\begin{figure}
\centering
\includegraphics[width=0.88\linewidth]{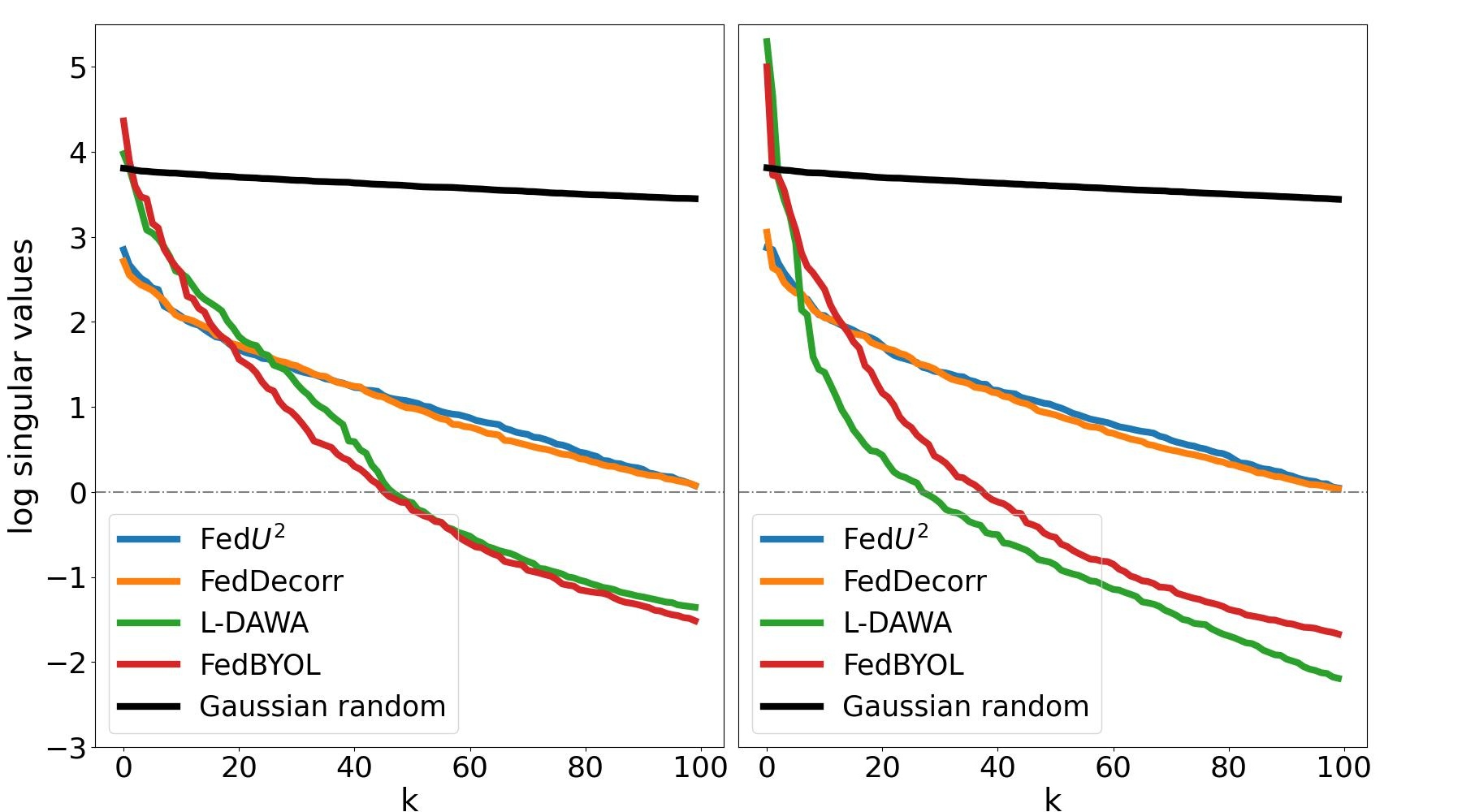}
% \vspace{-0.4cm}
\caption{Top k log singular values of the covariance matrix of global model (left) and local model (right) representations.}
\label{fig:SVD_local}
\end{figure}

\begin{figure*}[t]
\centering
\subfloat[FedBYOL]
{
    \includegraphics[scale=.11]{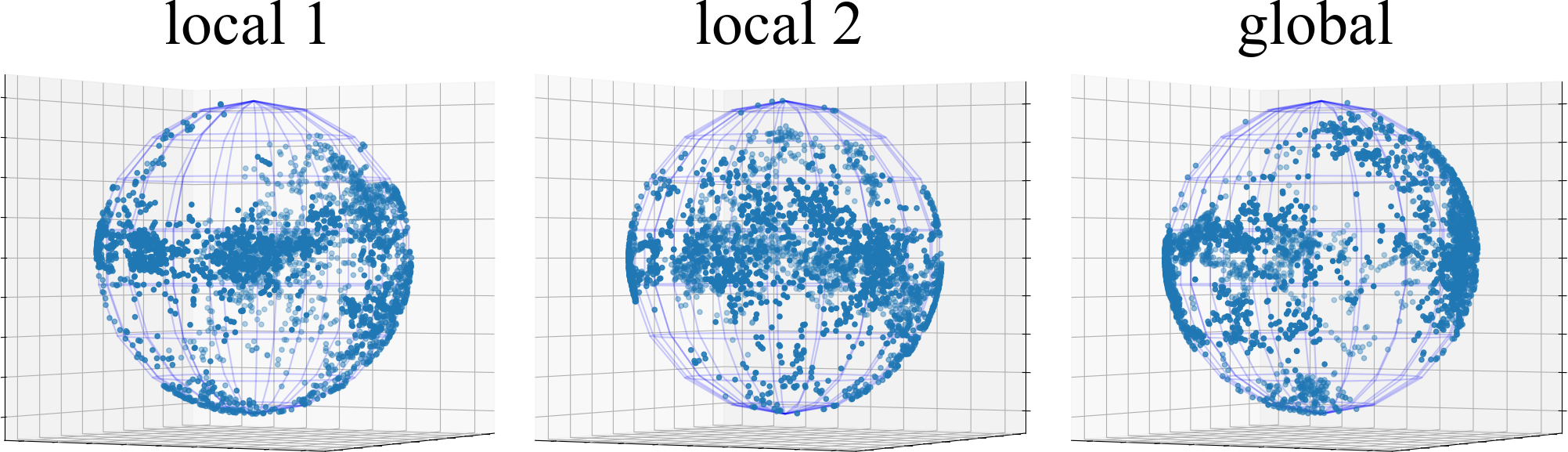}
    \label{fig:collapse_fedbyol}
}
\subfloat[FedDecorr]
{
    \includegraphics[scale=.11]{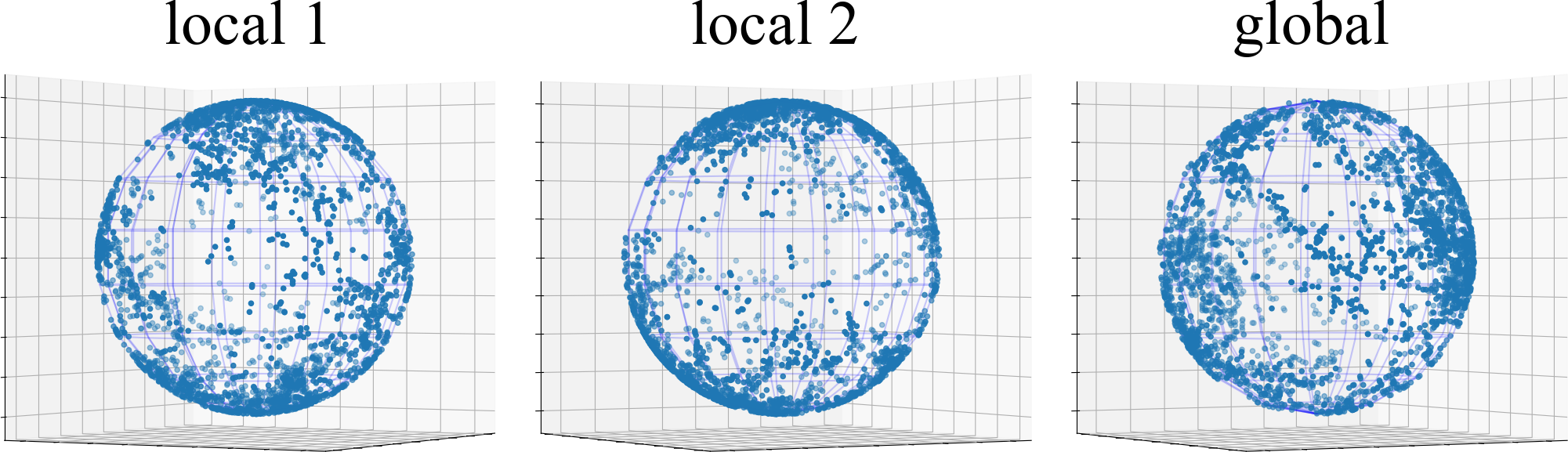}
    \label{fig:collapse_fedDecorr}
}\\
\subfloat[L-DAWA]
{
    \includegraphics[scale=.11]{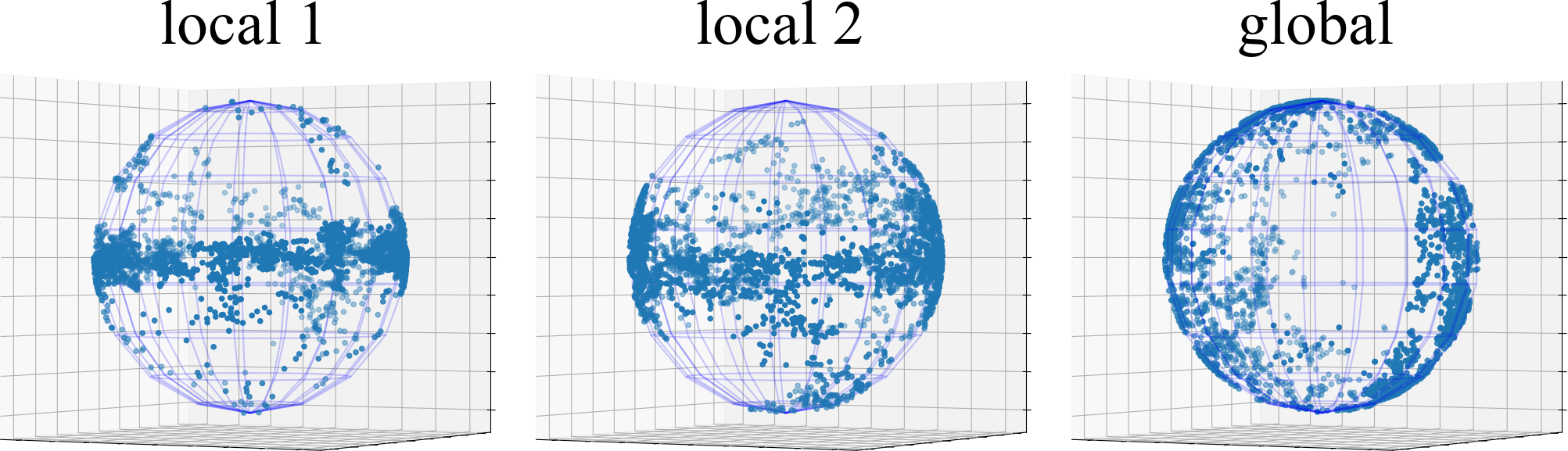}
    \label{fig:collapse_ldawa}
}
\subfloat[\modelname]
{
    \includegraphics[scale=.11]{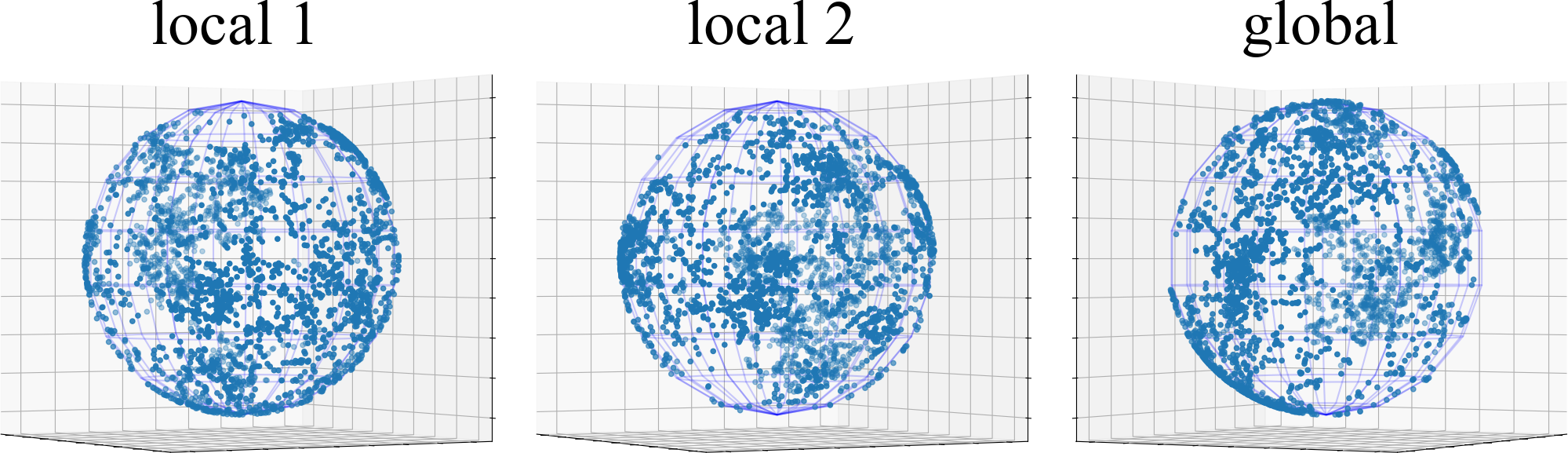}
    \label{fig:collapse_feduu}
}
\caption{The representations collapse issue on the sphere using BYOL model (on CIFAR10 $\alpha=0.1$ Cross-silo). The more blank representation space means the more severe collapse issue is.}
\label{fig:collapse}
\end{figure*}

\subsection{Representation Visualization}
\nosection{Analysis of Representation Collapse Entanglement}
To study the representation collapse entanglement caused by non-IID data, we capture the representation covariance matrices of $N$ test data points on CIFAR10 from both the global and local BYOL models of \problem~methods, i.e., FedDecorr, L-DAWA, FedBYOL, and \modelname.
And we utilize the singular value decomposition on each of the representation covariance matrices, and visualize the top-100 singular values in Fig.~\ref{fig:SVD_local}.
Both L-DAWA and FedBYOL suffer from severe representation collapse, because they have less 
 singular values beyond 0 than FedDecorr and \modelname.
The representation collapses in global model and local model are consistent, proving that collapse impacts are entangled intricately.
% 错中复杂的
% 
Compared with the singular values decomposed from the covariance matrix of Gaussian random samples, the singular values of FedDecorr and \modelname are not similar.
Because a fully uniform distribution breaks down the alignment effect and deteriorates clustering.
Furthermore, in Fig.~\ref{fig:collapse}, we visualize the representation collapse on 3-D spherical space, where the existing \problem~methods leave evident blank space and suffer from collapse entanglements.

\begin{figure}[t]
\centering
\subfloat{
\centering
\includegraphics[width=\linewidth]{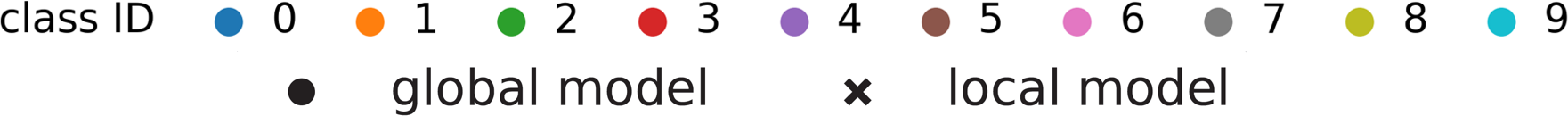} %
}\\
\subfloat[FedBYOL]{
\centering
\includegraphics[width=0.48\linewidth]{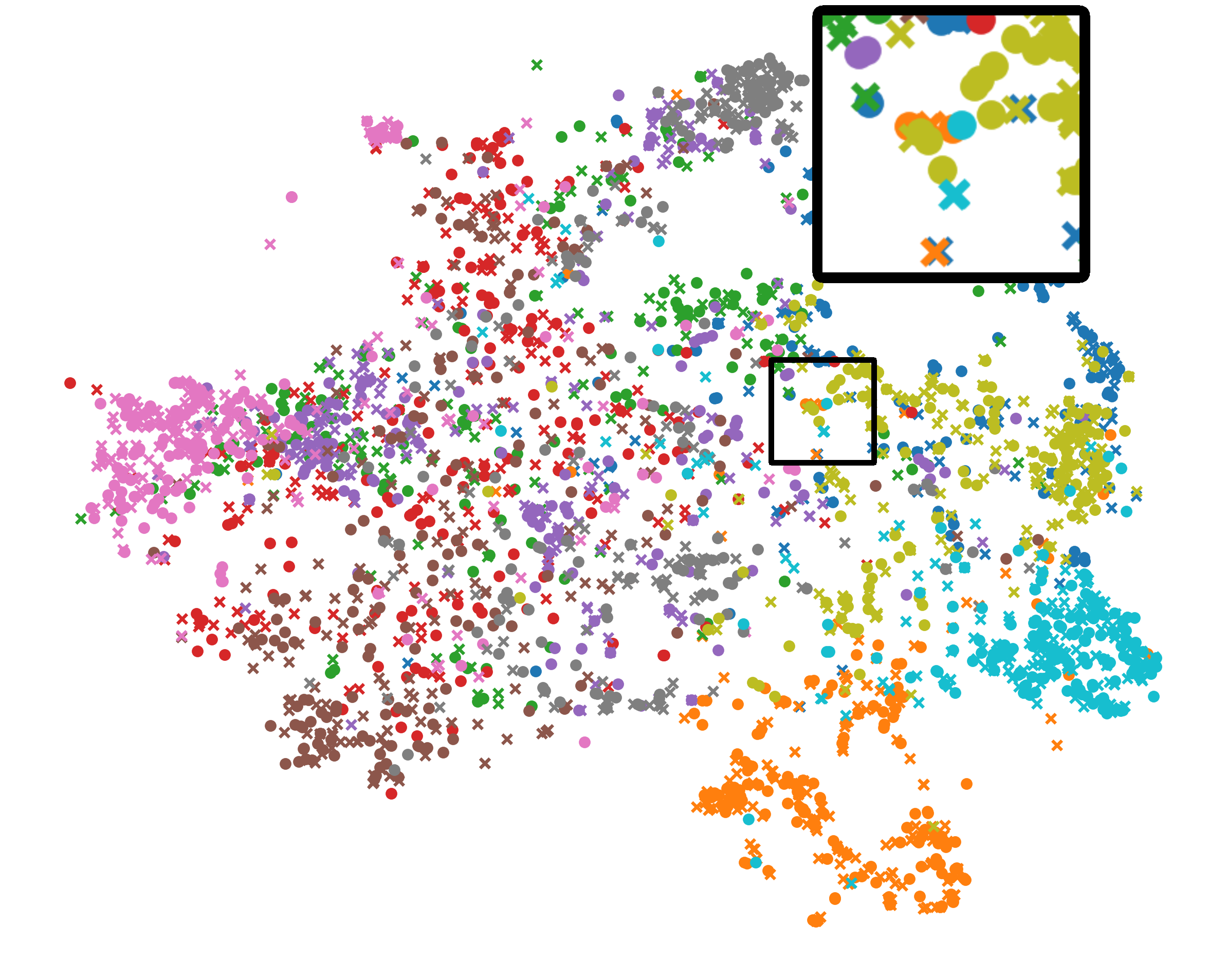} %
}
\subfloat[FedDecorr]{
\centering
\includegraphics[width=0.48\linewidth]{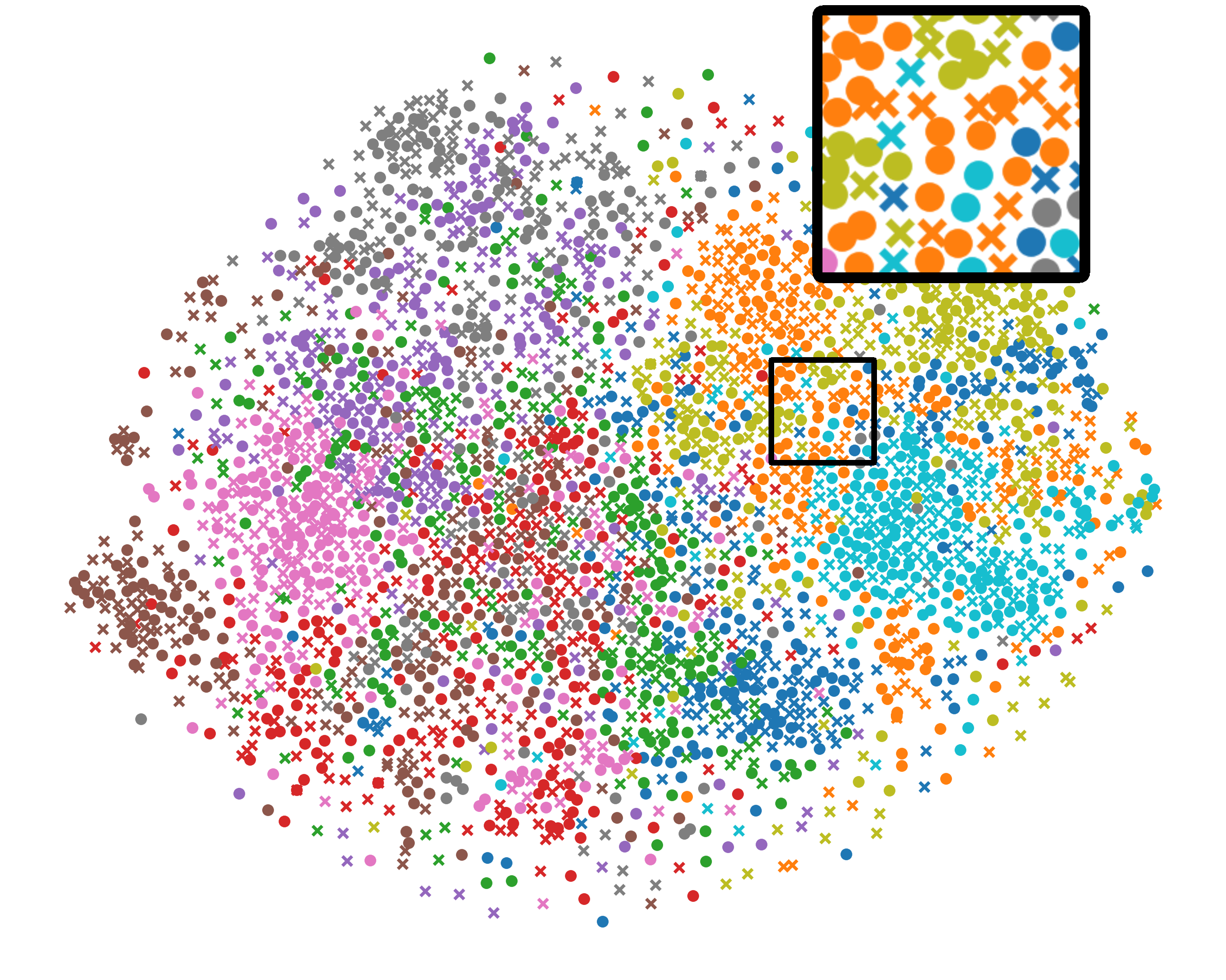} %
}\\
\subfloat[L-DAWA]{
\centering
\includegraphics[width=0.48\linewidth]{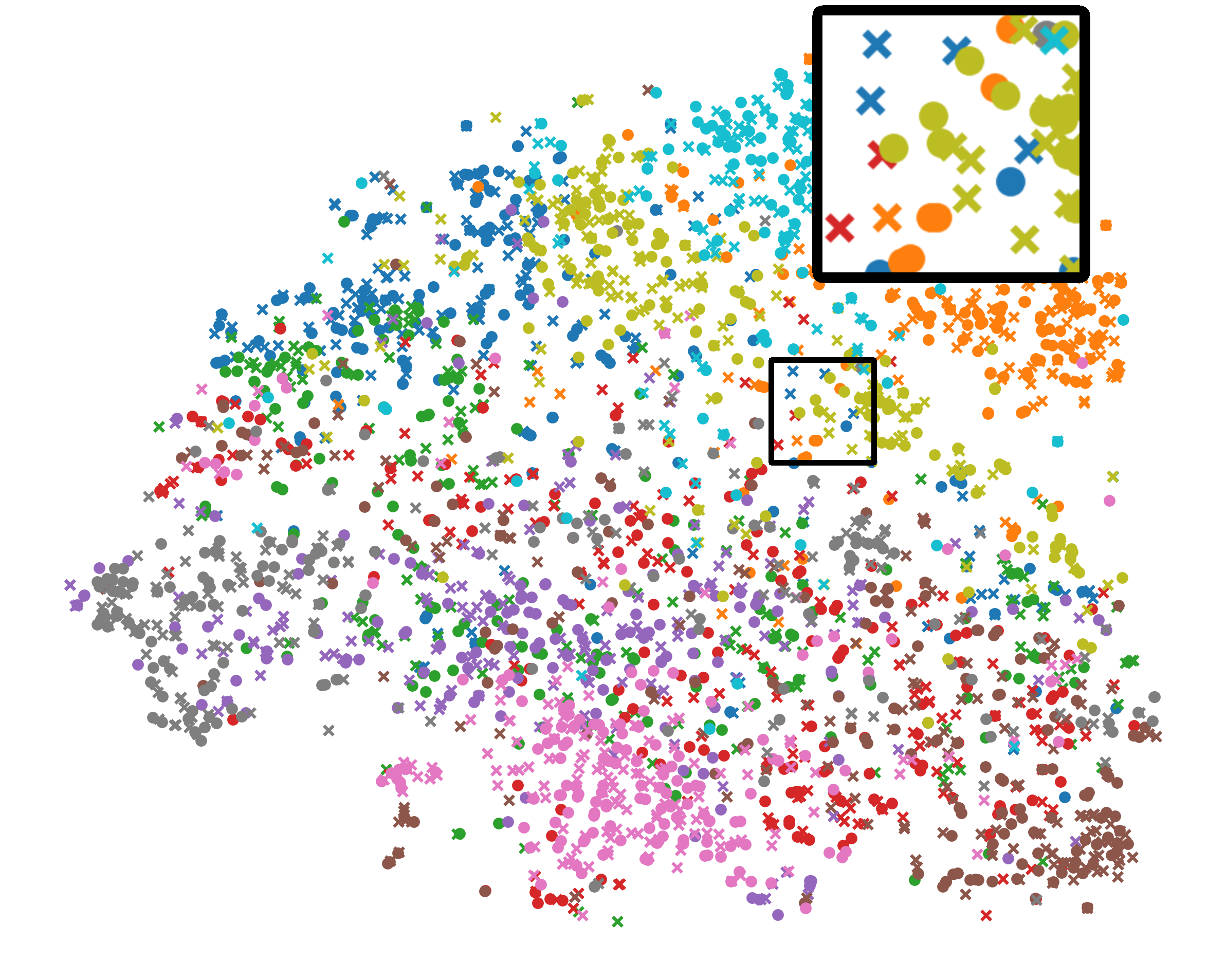} % 
}
\subfloat[\modelname]{
\centering
\includegraphics[width=0.48\linewidth]{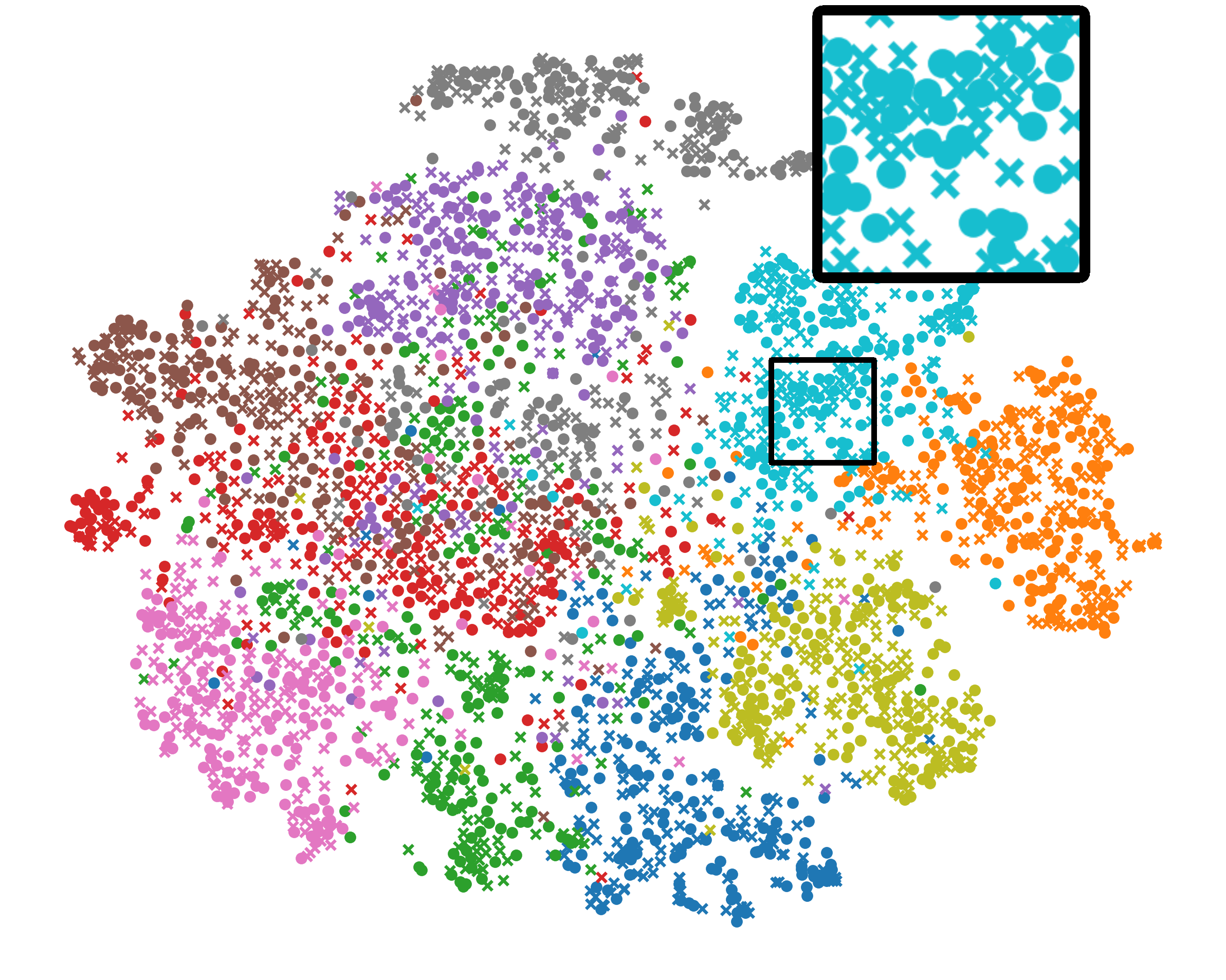} %
}
\caption{The distributions of 
data representations 
using global and local BYOL model (on  CIFAR10 $\alpha=0.1$ Cross-silo).}
\label{fig:unified}
\end{figure}

\nosection{Analysis of Unified Representation}
We also use t-SNE~\cite{tsne} to picture the 2-D representation of both global (circle) and local (cross) BYOL models in Fig.~\ref{fig:unified}.
There are three interesting conclusions:
Firstly, \modelname~has clearer cluster boundary than FedDecorr, validating that directly decomposing the Frobenius norm of representations deteriorates generalization.
Secondly, compared with FeBYOL, the global and local representations of L-DAWA are looser, implying its ineffectiveness in controlling conflicting model deviations.
Lastly, with the effect of \moduleB, \modelname~achieves tighter distribution consistency between global and local representations, as well as more clear cluster bound.

\begin{figure}[t]
\centering
\subfloat[BYOL]{
\centering
\includegraphics[width=0.48\linewidth]{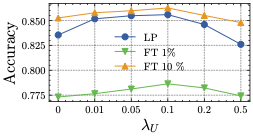} % 
}
\subfloat[Simsiam]{
\centering
\includegraphics[width=0.48\linewidth]{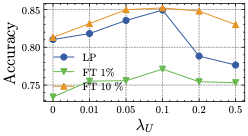} %
}
\caption{The effect of $\lambda_U$ (on  CIFAR10 $\alpha=0.1$ Cross-silo).}
\label{fig:lambda_U}
\end{figure}

\nosection{Hyper-parameters sensitivity}
We consider the sensitivity of highly relevant hyper-parameters, i.e., the effect of uniformity term $\lambda_U=\{0, 0.01, 0.05,0.1, 0.2, 0.5\}$ on Cifar10 Cross-silo ($\alpha=0.1)$, in Fig.~\ref{fig:lambda_U}.
We set $\lambda_U=0.1$ in experiments since it reaches the highest performance.
And we leave the number of clients $K= \{5, 10, 20,50,100\}$ and the local epochs $E= \{5, 10,20, 50\}$ in Appendix B.
% 
% Fig.~\ref{fig:hyperparameter_E},

\section{Conclusion}
In this work,
% we rethink two coupling challenges of representation in federated unsupervised learning (\problem), i.e., mitigating representation collapse entanglement and obtaining unified representation spaces.
% % 
% To tackle the above challenges,
we propose a \problem~framework, i.e., \modelname, to enhance \textbf{U}niform and \textbf{U}nified representation.
\modelname~consists of flexible uniform regularizer (\moduleA) and efficient unified aggregator (\moduleB).
\moduleA~encourages data representations to uniformly distribute in a spherical Gaussian space, mitigating representation collapse and its subsequent entangled impacts.
\moduleB~further constrains the consistent optimization improvements among different client models, which is good for unified representation.
 In our empirical studies, we set both cross-silo and cross-device settings, and  
 conduct experiments on CIFAR10 and CIFAR100 datasets, which extensively validate the superiority of \modelname.

\nosection{Acknowledgements} 
% This work was supported in part by the National Key
% R\&D Program of China (No. 2022YFF0902704), the Ten
% Thousand Talents Program of Zhejiang Province for Leading
% Experts (No. 2021R52001), and Ant Group.
This work was supported by National Key R\&D Program of China
(2022YFB4501500, 2022YFB4501504), and the National Natural Science Foundation of China (No.72192823).

% This work is supported by National Key R&D Program of China
% (2022YFB4501500, 2022YFB4501504).

%%%%%%%%% REFERENCES
{\small
\bibliographystyle{ieee_fullname}
% \balance
\bibliography{egbib}
}

\clearpage

\appendix

In the supplemental materials, we provide the theoretical analysis in Appendix A, and the additional experimental details in Appendix B.

\section{Theoretical Analysis}
\subsection{Optimization Consistency of Model Deviation}
In the following, we provide the theorem related to optimization consistency of model deviations.
When the aggregating weights, i.e., $\boldsymbol{p}$, achieve optimal, the model deviation rate is equally contributed to global updating.

\begin{theorem}[Optimization consistency of model deviations]
Rethinking the Lagrangian of dual form in Eq.~\eqref{eq:mtl_cc_dual}, 
\begin{equation}
\boldsymbol{J} =\min _{\boldsymbol{p}} \frac{\eta^2_g}{2\phi} \|{(\nabla\log  {\boldsymbol{u}})}^\top \boldsymbol{p}\|^2 + \lambda_E \boldsymbol{p}^\top \boldsymbol{1},
\label{supp_eq:damo_dual_add}
\end{equation}
it holds $\nabla \log( {u}_i\left(\boldsymbol{\theta}^{t}_g\right)) =\nabla \log( {u}_j\left(\boldsymbol{\theta}^{t}_g\right))$,  $\forall i \neq j\in [K]$.
\begin{proof}

By deviating Eq.~\eqref{supp_eq:damo_dual_add} with regarding to $\boldsymbol{p}$, we can obtain $ {(\nabla\log  {\boldsymbol{u}})}  \nabla {\log({\boldsymbol{u}})}^\top \boldsymbol{p}^{*}=-\frac{\phi\lambda_E}{\eta_g^2}\boldsymbol{I}$.
Remind that $\boldsymbol{d}^{*}= \frac{\eta_g}{\phi} {(\nabla\log  {\boldsymbol{u}})}^\top \boldsymbol{p}^{*}$, 
for client $i\neq j \in [K]$ and ${\eta_g}\rightarrow 0$, we finally have consistent model deviation change rate as below:
\begin{equation}
\begin{aligned}
\quad \lim _{\eta_g \rightarrow 0} c_i\left(\eta_g, \boldsymbol{d}^*\right)=\nabla \log( {u}_i\left(\boldsymbol{\theta}^{*}_g\right))\boldsymbol{d}^* \\
=\nabla \log( {u}_j\left(\boldsymbol{\theta}^{*}_g\right)) \boldsymbol{d}^*=\lim _{\eta_g \rightarrow 0} c_j\left(\eta_g, \boldsymbol{d}^*\right).
\end{aligned}
\end{equation}
\end{proof}
\label{supp_th:equal_deviation}
\end{theorem}
Therefore, the global model updates with a direction that balances all model deviation change rates, obtaining consistent parameters for server and client models.

\subsection{Bound of Client Model Divergence}
In this part, we first introduce mild and general assumptions~\cite{zhihua_convergence}, and induct the model updating divergence bound for each client.

\begin{assumption}
\label{supp_asu:l-lip}
Let $F_k(\boldsymbol{\theta})$ be the expected model objective for client $k$, and assume
    $F_1, \cdots, F_K$ are all L-smooth, i.e., for all $\boldsymbol{\theta}_k$, $F_k(\boldsymbol{\theta}_k) \leq F_k(\boldsymbol{\theta}_k)+(\boldsymbol{\theta}_k- \boldsymbol{\theta}_k)^\top \nabla F_k(\boldsymbol{\theta}_k)+\frac{L}{2}\|\boldsymbol{\theta}_k- \boldsymbol{\theta}_k\|_2^2$.
\end{assumption}
\begin{assumption}
\label{supp_asu:cvx}
   Let $F_1, \cdots, F_N$ are all $\mu$-strongly convex: for all $\boldsymbol{\theta}_k$, $F_k(\boldsymbol{\theta}_k) \geq F_k(\boldsymbol{\theta}_k)+(\boldsymbol{\theta}_k- \boldsymbol{\theta}_k)^\top \nabla F_k(\boldsymbol{\theta}_k)+\frac{\mu}{2}\|\boldsymbol{\theta}_k- \boldsymbol{\theta}_k\|_2^2$.
\end{assumption}

\begin{assumption}
\label{supp_asu:bound_var}
Let $\xi^t_k$ be sampled from the $k$-th client's local data uniformly at random. The variance of stochastic gradients in each client is bounded: $\mathbb{E}\left\|\nabla F_k\left(\boldsymbol{\theta}_k^{t}, \xi^t_k\right)-\nabla F_k\left(\boldsymbol{\theta}_k^{t}\right)\right\|^2 \leq \sigma_k^2$. % $(\forall k=1, \cdots, K)$.
\end{assumption}

\begin{assumption}
\label{supp_asu:bound_grad_norm}
The expected squared norm of stochastic gradients is uniformly bounded, i.e., $\mathbb{E}\left\|\nabla F_k\left(\boldsymbol{\theta}_k^{t}, \xi^t_k\right)\right\|^2 \leq V^2$ for all $k=1, \cdots, N$ and $t=1, \cdots, T-1$
\end{assumption}

 Next, we introduce the lemma related to the bound of client model divergence.
\begin{lemma}[Bound of Client Model Divergence]
\label{supp_lemma:div}
With assumption~\ref{supp_asu:bound_grad_norm}, $\eta_t$ is non-increasing and $\eta_t< 2\eta_{t+E}$ (learning rate of t-th round and E-th epoch) for all $t\geq 0$, there exists $t_0 \leq t$, such that $t-t_0\leq E-1$ and $\boldsymbol{\theta}^{t_0}_k= \boldsymbol{\theta}^{t_0}$ for all $k\in [N]$. It follows that 
\begin{equation}
\label{supp_eq:var_bound}
\mathbb{E} \left[ \sum_k^K p_k \|\boldsymbol{\theta}^{t}- \boldsymbol{\theta}^{t}_k\|^2 \right]\leq 4\eta_t^2 {(E-1)}^2 V^2.
\end{equation}
\begin{proof}
Let $E$ be the maximal local epoch.
For any round $t>0$, communication rounds from $t$ to $t_0$ exist $t-t_0 < E-1$.
and the global model $\boldsymbol{\theta}^{t_0}$ and each local model $\boldsymbol{\theta}_k^{t_0}$ are same at round $t_0$.

\begin{equation}
\small
\begin{aligned}
&\mathbb{E} \left[ \sum_k^K p_k \|\boldsymbol{\theta}^{t}- \boldsymbol{\theta}^{t}_k\|^2 \right]\\
&= \mathbb{E} \left[ \sum_k^K p_k \|(\boldsymbol{\theta}^{t}_k-\boldsymbol{\theta}^{t_0})-(\boldsymbol{\theta}^{t}-\boldsymbol{\theta}^{t_0})\|^2 \right] &(17a)\\
&\leq \mathbb{E}\sum_k^K p_k \|\boldsymbol{\theta}^{t}_k-\boldsymbol{\theta}^{t_0}\|^2 &(17b)\\
&=  \mathbb{E}\sum_k^K p_k \left\|\sum_{t=t0}^{t-1}\eta_t \nabla F_k(\boldsymbol{\theta}^t_k, \xi_k^t)\right\|^2 &(17c)\\
&\leq \mathbb{E}\sum_k^K p_k (t-t_0)\sum_{t=t0}^{t-1} \eta_{t_0}^2 \left\|\nabla F_k(\boldsymbol{\theta}_t^k, \xi_k^t)\right\|^2 &(17d)\\
&\leq 4\eta^2_{t} (E-1)^2V^2,&(17e)
\end{aligned}
\nonumber
\end{equation}
where the Eq.~(17b) holds since $\mathbb{E} (\boldsymbol{\theta}^{t}_k-\boldsymbol{\theta}^{t_0})=\boldsymbol{\theta}^{t}-\boldsymbol{\theta}^{t_0}$, and $\mathbb{E}\|X-\mathbb{E}(X)\|\leq \mathbb{E}\|X\|$, and Eq.~(17d) derives from Jensen inequality.
\end{proof}
\end{lemma}

\subsection{Convergence Error Bound}

\begin{definition}[Heterogeneity Quantification~\cite{zhihua_convergence}]
Let $F^*$ and $F_k^*$ be the minimum values of $F$ and $F_k$, respectively. 
We use the term $\Gamma=F^*-\sum_{k=1}^N p_k F_k^*$ for quantifying the degree of non-IID.
If the data are IID, then $\Gamma$ obviously goes to zero as the number of samples grows. 
If the data are non-IID, then $\Gamma$ is nonzero, and its magnitude reflects the heterogeneity of the data distribution.
\end{definition}

\begin{theorem}[Convergence Error Bound]
\label{supp_th:convergence}
Let assumptions \ref{supp_asu:l-lip}-\ref{supp_asu:bound_grad_norm} hold, and $L, \mu, \sigma_k, V$ be defined therein.
Let $\kappa=\frac{L}{\mu}, \gamma=\max\{8\kappa, E\}$ and the learning rate $\eta_t=\frac{2}{\mu(\gamma+t)}$.
The \modelname~with full client participation satisfies 
$$
\mathbb{E}\left[F\left({\overline{\boldsymbol{\theta}}}^t\right)\right]-F^* \leq \frac{\kappa}{\gamma+t}\left(\frac{2 B}{\mu}+\frac{\mu(\gamma+1)}{2} \|\boldsymbol{\theta}^t - \boldsymbol{\theta}^{*}\|^2\right),
$$
where $B=4(E-1)^2V^2 +K+2\Gamma$.
\begin{proof}
By L-smooth assumption~\ref{supp_asu:l-lip}, we can obtain:
\begin{equation}
\label{supp_eq:obj_bound}
\begin{aligned}
    &\mathbb{E} \left[F(\boldsymbol{\theta}^t) -F(\boldsymbol{\theta}^{*})\right] \\
    &\leq \mathbb{E} \left[(\boldsymbol{\theta}^t-\boldsymbol{\theta}^{*})^{\top} \nabla F\left(\boldsymbol{\theta}^{*}\right)+\frac{L}{2}\left\|\boldsymbol{\theta}^t-\boldsymbol{\theta}^{*}\right\|^2 \right]\\
    & =\mathbb{E} \left[ \frac{L}{2}\left\|\boldsymbol{\theta}^t-\boldsymbol{\theta}^{*}\right\|^2 \right].
\end{aligned}
\end{equation}

Since the updating in \moduleB~is $\boldsymbol{\theta}^{t+1}= \boldsymbol{\theta}^t - \eta_t \boldsymbol{d}^t$ 
for $\boldsymbol{d}^{*}_{t}= \frac{\eta_t}{\phi} {(\nabla\log  {\boldsymbol{u}}^t)}^\top \boldsymbol{p}^{*}_{t} =\sum_k^K p_k \nabla F_k(\boldsymbol{\theta}^t_k)$,
we can rewrite it as:
\begin{equation}
\label{supp_eq:model_updating}
\small
\begin{aligned}
&\|\boldsymbol{\theta}^{t+1}- \boldsymbol{\theta}^{*}\|^2\\
&= \|\boldsymbol{\theta}^{t} - \eta_t\boldsymbol{d}_t -\boldsymbol{\theta}^{*}\|^2\\
& =\|\boldsymbol{\theta}^{t}- \boldsymbol{\theta}^{*}\|^2 -2\left<\boldsymbol{\theta}^{t}- \boldsymbol{\theta}^{*},  \eta_t \boldsymbol{d}_t\right> +\eta\|\boldsymbol{d}_t\|^2.\\
% & \leq \|\boldsymbol{\theta}^{t}- \boldsymbol{\theta}^{*}\|^2 +\eta\|\boldsymbol{d}_t\|^2,\\
\end{aligned}
\end{equation}
% \begin{equation}
% \label{supp_eq:model_updating}
% \small
% \begin{aligned}
% &\|\boldsymbol{\theta}^{t+1}- \boldsymbol{\theta}^{*}\|^2\\
% &= \|\boldsymbol{\theta}^{t} - \eta_t {(\nabla\log  {\boldsymbol{u}}^t)}^\top \boldsymbol{p}- \boldsymbol{\theta}^{*}\|^2\\
% & =\|\boldsymbol{\theta}^{t}- \boldsymbol{\theta}^{*}\|^2 -2\left<\boldsymbol{\theta}^{t}- \boldsymbol{\theta}^{*},  \eta_t {(\nabla\log  {\boldsymbol{u}}^t)}^\top\boldsymbol{p}\right> +\eta\|{(\nabla\log  {\boldsymbol{u}}^t)}^\top\boldsymbol{p}\|^2,\\
% & \leq \|\boldsymbol{\theta}^{t}- \boldsymbol{\theta}^{*}\|^2 +\eta\|{(\nabla\log  {\boldsymbol{u}}^t)}^\top\boldsymbol{p}\|^2,\\
% \end{aligned}
% \end{equation}
% where $\eta= \frac{\eta^2_t}{\phi}$.

Next, we induce the bound of the second term.
\begin{equation}
\small
\begin{aligned}
&\left<\boldsymbol{\theta}^{t}- \boldsymbol{\theta}^{*},  \eta_t \boldsymbol{d}_t\right>\\
% &=\left<\boldsymbol{\theta}^{t}-\boldsymbol{p}^\top \boldsymbol{\theta}^t_k-  \boldsymbol{\theta}^{*} +\boldsymbol{p}^\top \boldsymbol{\theta}^t_k,  \eta_t \boldsymbol{d}_t\right>\\
% &=\left<\boldsymbol{\theta}^{t}-\boldsymbol{p}^\top \boldsymbol{\theta}^t_k,  \eta_t \boldsymbol{d}_t\right>-\left<\boldsymbol{\theta}^{*}- \boldsymbol{p}^\top\boldsymbol{\theta}^t_k,  \eta_t \boldsymbol{d}_t\right>\\
& =\sum_k^K p_k \eta_t \left<\boldsymbol{\theta}^t- \boldsymbol{\theta}^t_k, \nabla F_k(\boldsymbol{\theta}_k^t)\right>- \sum_k^K p_k \eta_t \left<\boldsymbol{\theta}^t_k- \boldsymbol{\theta}^{*}, \nabla F_k(\boldsymbol{\theta}_k^t)\right>
\end{aligned}
\label{supp_eq:term_of_expansion}
\end{equation}
\small
By Cauchy-Schwarz inequality and AM-GM inequality, we have inequality of the first term of Eq.~\eqref{supp_eq:term_of_expansion}:
\begin{equation}
\label{supp_eq:cs_ieq}
-2\left<\boldsymbol{\theta}^t- \boldsymbol{\theta}^t_k, \nabla F_k(\boldsymbol{\theta}_k^t)\right> \leq \frac{1}{\eta_t} \|\boldsymbol{\theta}^t- \boldsymbol{\theta}^t_k\|^2+\eta_t \|\nabla F_k(\boldsymbol{\theta}_k^t)\|^2.
\end{equation}
By the $\mu-$strong convexity of $F_k(\cdot)$, we have 
\begin{equation}
\small
\label{supp_eq:strong_cvx}
-\left<\boldsymbol{\theta}^t_k- \boldsymbol{\theta}^{*}, \nabla F_k(\boldsymbol{\theta}_k^t)\right> \leq -\left(F_k(\boldsymbol{\theta}^t_k)- F_k(\boldsymbol{\theta}^{*})\right) -\frac{\mu}{2} \|\boldsymbol{\theta}^t_k- \boldsymbol{\theta}^{*}\|^2.
\end{equation}
In Theorem~\ref{supp_th:equal_deviation}, we get $ {(\nabla\log  {\boldsymbol{u}})}^\top  \nabla {\log({\boldsymbol{u}})} \boldsymbol{p}^{*}=-\frac{\phi\lambda_E}{\eta_t^2}\boldsymbol{I}$, which indicates that:
\begin{equation}
\begin{aligned}
    \|\boldsymbol{d}_t\|^2& = \frac{\eta_t^2}{\phi} \|{(\nabla\log {\boldsymbol{u}}^t)}^\top\boldsymbol{p}\|^2\\
&=\lambda_E  \|\boldsymbol{p}^{\top}\|^2\\
& \leq K, 
\end{aligned}
\label{supp_eq:d_norm}
\end{equation}
where the last inequation holds due to $\lambda_E<1$ and $\|\boldsymbol{p}\|\leq K$.
By combining Eq.~\eqref{supp_eq:cs_ieq}-\eqref{supp_eq:d_norm} and Lemma ~\ref{supp_lemma:div}, it follows that 
\begin{equation}
\label{supp_eq:bound}
\small
\begin{aligned}
&\|\boldsymbol{\theta}^{t+1}- \boldsymbol{\theta}^{*}\|^2\\
&= \|\boldsymbol{\theta}^{t} - \eta_t \boldsymbol{d}_t- \boldsymbol{\theta}^{*}\|^2\\
&\leq \|\boldsymbol{\theta}^{t}- \boldsymbol{\theta}^{*}\|^2 +
\eta_t \sum_k^K p_k \left(\frac{1}{\eta_t}\|\boldsymbol{\theta}^{t}- \boldsymbol{\theta}^{t}_k\|^2+ \eta_t \|\nabla F_k(\boldsymbol{\theta}_k^t)\|^2 \right)\\
&+
2 \eta_t \sum_k^K p_k\left( -\left(F_k(\boldsymbol{\theta}^t_k)- F_k(\boldsymbol{\theta}^{*})\right) -\frac{\mu}{2} \|\boldsymbol{\theta}^t_k- \boldsymbol{\theta}^{*}\|^2\right)
+\eta_t^2 \|\boldsymbol{d}_t\|^2\\
&=(1-\mu\eta_t)\|\boldsymbol{\theta}^t- \boldsymbol{\theta}^{*}\|^2 + \sum_k^K p_k\|\boldsymbol{\theta}^{t}- \boldsymbol{\theta}^{t}_k\|^2 + \eta_t^2 K + 2\eta_t^2\Gamma \\
&\leq (1-\mu\eta_t)\|\boldsymbol{\theta}^t- \boldsymbol{\theta}^{*}\|^2 + 4\eta_t^2 (E-1)^2V^2 +\eta_t^2 K+2\eta_t^2\Gamma.
\end{aligned}
\end{equation}

% Note that GNE holds the optimization, i.e., $\boldsymbol{G}^\top \boldsymbol{G}\boldsymbol{p}=\frac{1}{\boldsymbol{p}}$, which guarantees the last term of right hand is a constant, i.e., $\|\boldsymbol{G}\boldsymbol{p}\|^2= {\boldsymbol{p}}^\top {\boldsymbol{G}}^\top \boldsymbol{G}\boldsymbol{p} =K.$
% % 
% Next, we can expand 
% \begin{equation}
% \label{supp_eq:Gp}
% \begin{aligned}
%     \boldsymbol{G}\boldsymbol{p} &= \sum_{k=1}^{K} p_k \Delta_{\theta_k}\\
% &= \sum_{k=1}^{K} p_k (\theta_k^{t+1}-\theta^{t})\\
% &= \sum_{k=1}^{K} p_k (\theta_k^{t+1, e=E}-\theta_k^{t+1,e=0})\\
% &= \sum_{k=1}^{K} p_k \nabla F_k(\theta_k^t),
% \end{aligned}
% \end{equation}
% where $e$ means the epochs executed.

Lastly, let $D_t = \mathbb{E} \|\boldsymbol{\theta}^t- \boldsymbol{\theta}^{*}\|^2$,
it follows that 
\begin{equation}
\label{supp_eq:D}
D_{t+1}\leq (1-\eta_t \mu)D_t +\eta_t^2 B,
\end{equation}
where $B=4(E-1)^2V^2 +K+2\Gamma$.

For a diminishing stepsize, $\eta_t = \frac{\beta}{t+\gamma}$ for some $\beta>\frac{1}{\mu}$ and $\gamma>0$ such that $\eta_1\leq \min \{\frac{1}{\mu}, \frac{1}{4L}\}=\frac{1}{4L}$ and $\eta_t\leq 2\eta_{t+E}$.
For $v=\max\{\frac{\beta^2B}{\beta \mu -1}, (\gamma+1)D_1\}$, by definition, it holds $D_t \leq \frac{v}{\gamma+t}$ for $t=1$.
Assume $D_t \leq \frac{v}{\gamma+t}$ holds, then we expand as below:
\begin{equation}
\begin{aligned}
D_{t+1} & \leq\left(1-\eta_t \mu\right) D_t+\eta_t^2 B \\
& \leq\left(1-\frac{\beta \mu}{t+\gamma}\right) \frac{v}{t+\gamma}+\frac{\beta^2 B}{(t+\gamma)^2} \\
& =\frac{t+\gamma-1}{(t+\gamma)^2} v+\left[\frac{\beta^2 B}{(t+\gamma)^2}-\frac{\beta \mu-1}{(t+\gamma)^2} v\right] \\
& \leq \frac{v}{t+\gamma+1} .
\end{aligned}
\end{equation}
Recall Eq.~\eqref{supp_eq:obj_bound}, we finally catch:
\begin{equation}
    \mathbb{E} \left[F(\theta^t) -F(\theta^{*})\right] \leq \frac{L}{2} D_t\leq\frac{L}{2} \frac{v}{\gamma+t}.
\end{equation}

Following the specific case of~\cite{zhihua_convergence}, we can  
choose $\beta=\frac{2}{\mu}, \gamma=\max \left\{8 \frac{L}{\mu}, E\right\}-1$ and denote $\kappa=\frac{L}{\mu}$, then $\eta_t=\frac{2}{\mu} \frac{1}{\gamma+t}$.
One can verify that the choice of $\eta_t$ satisfies $\eta_t \leq 2 \eta_{t+E}$ for $t \geq 1$.
Then, we have
\begin{equation}
\begin{aligned}
v&=\max \left\{\frac{\beta^2 B}{\beta \mu-1},(\gamma+1) \Delta_1\right\}\\
&\leq \frac{\beta^2 B}{\beta \mu-1}+(\gamma+1) \Delta_1 \\
&\leq \frac{4 B}{\mu^2}+(\gamma+1) D_1
\end{aligned}
\end{equation}
and
\begin{equation}
\mathbb{E}\left[F\left({\overline{\boldsymbol{\theta}}}^t\right)\right]-F^* \leq \frac{L}{2} \frac{v}{\gamma+t} \leq \frac{\kappa}{\gamma+t}\left(\frac{2 B}{\mu}+\frac{\mu(\gamma+1)}{2} D_1\right).
\end{equation}
As we can see, \modelname~similarly converges to a generalization error bound as the FedAvg-like FL model with non-IID data. 
Discriminatively, benefiting from the optimization of \moduleB, the communication round multiplies with a smaller $B$. 
\end{proof}
\end{theorem}

\section{Experimental Supplementary}
\subsection{Hyper-parameter Sensitivity Analysis}
\begin{figure}[t]
\centering
\includegraphics[width=0.98\linewidth]{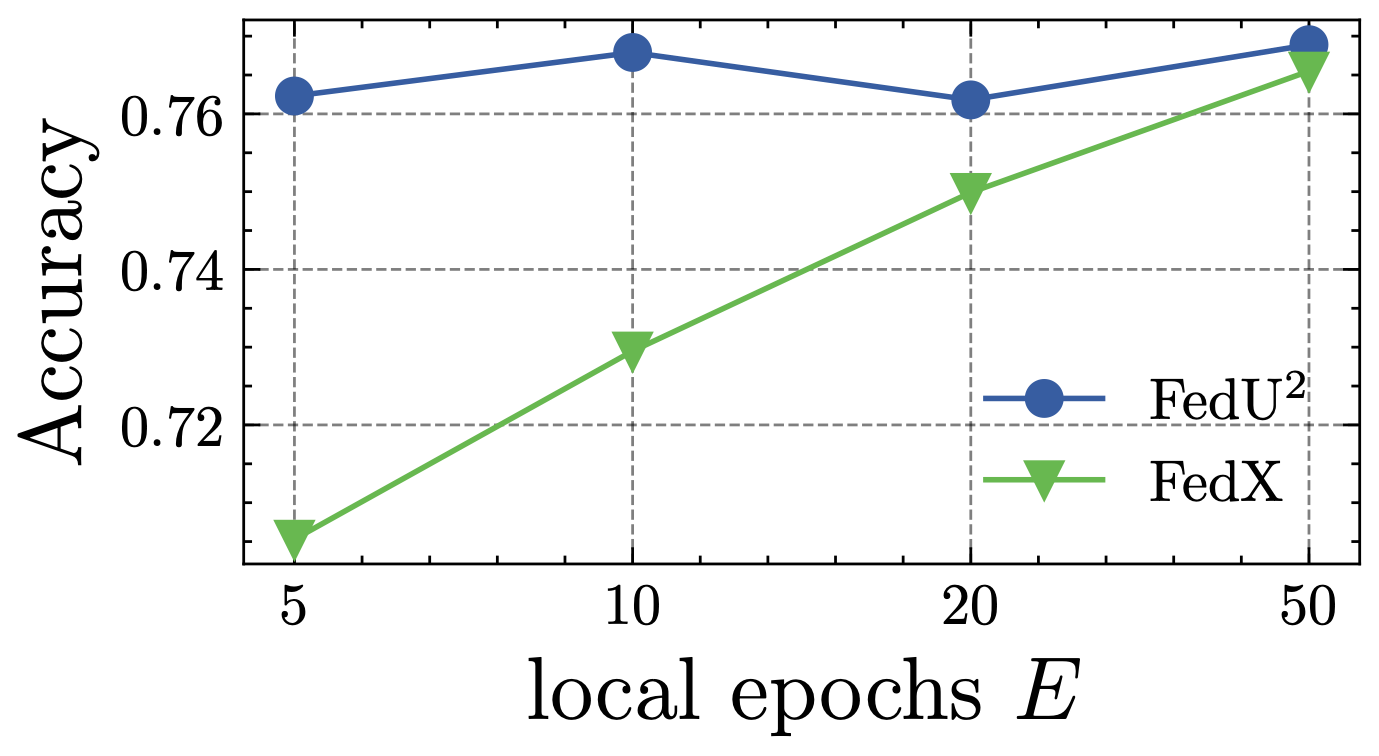}
% \vspace{-0.4cm}
\caption{The effect of local epochs $E$ (on CIFAR10 $\alpha=0.1$).}
\label{supp_fig:E}
\end{figure}

\begin{figure}[t]
\centering
\includegraphics[width=0.98\linewidth]{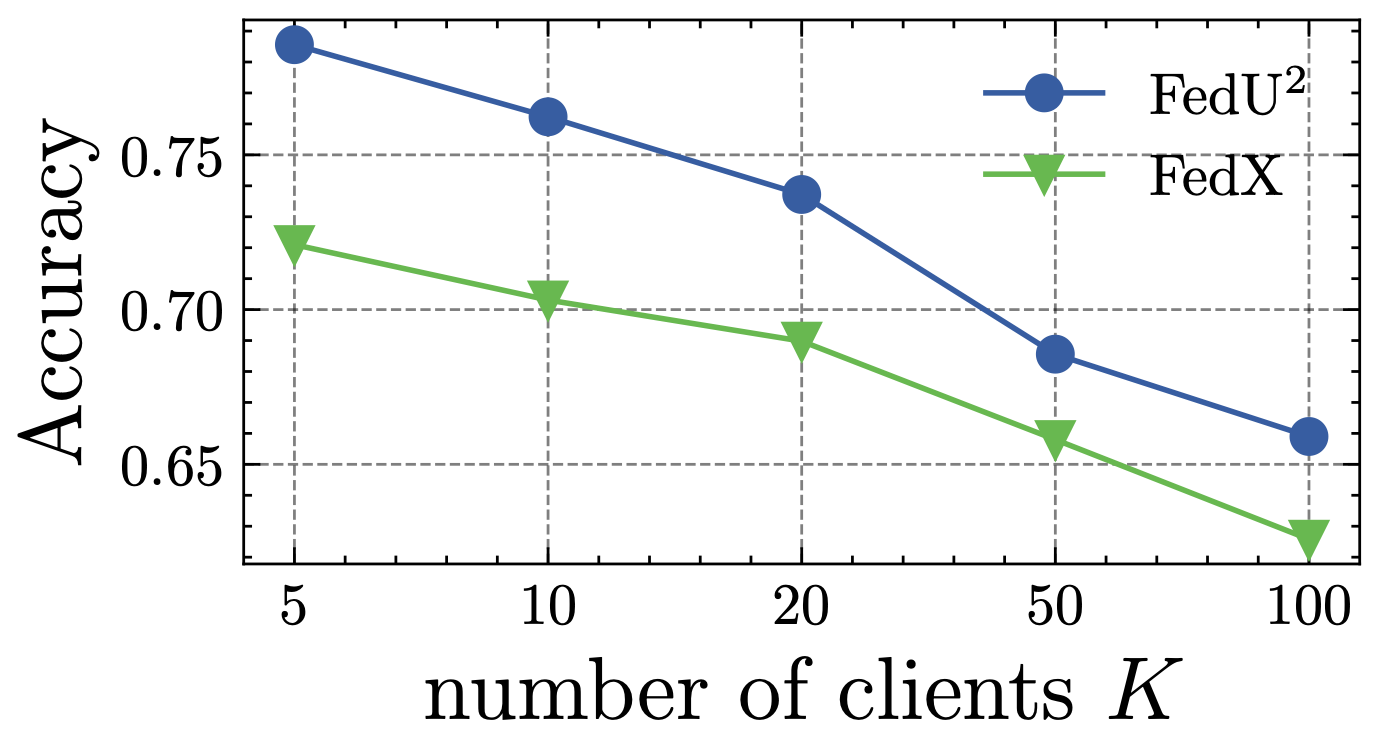}
% \vspace{-0.4cm}
\caption{The effect of client number $K$ (on CIFAR10 $\alpha=0.1$).}
\label{supp_fig:K}
\end{figure}

% We consider the sensitivity of highly relevant hyper-parameters, i.e., the effect of uniformity term in Fig.~\ref{fig:lambda_U} in the main paper.
% 
In the following, we study the sensitivity of remaining highly relevant hyper-parameters, i.e., the effect of client numbers and local epochs.
Specifically, we compare \modelname-SimCLR and its runner-up method, i.e., FedX-SimCLR, on CIFAR10 $\alpha=0.1$, by varying the local epochs $E= \{5, 10,20, 50\}$ in Fig.~\ref{supp_fig:E} and
the number of clients $K= \{5, 10, 20,50,100\}$ in Fig.~\ref{supp_fig:K}.
We train all models until converge to obtain fairly comparable results.
As we can see:
(1) With the increase of local epochs, each client of FedX-SimCLR obtains a better-performing model, while each client of \modelname-SimCLR is insensitive.
This states that \modelname~balances the client model deviation change rate in \moduleB, bringing the benefits of quick convergence.
(2) The performance of all methods decreases when the number of clients increases, but \modelname-SimCLR~consistently outperforms FedX-SimCLR.
It validates that enhancing uniform and unified representations will make \problem~methods more generalizable to the cases of various participants amounts.

\subsection{Enlarged Figures in Visualization}
\begin{figure*}[t]
\centering
\includegraphics[width=\linewidth]{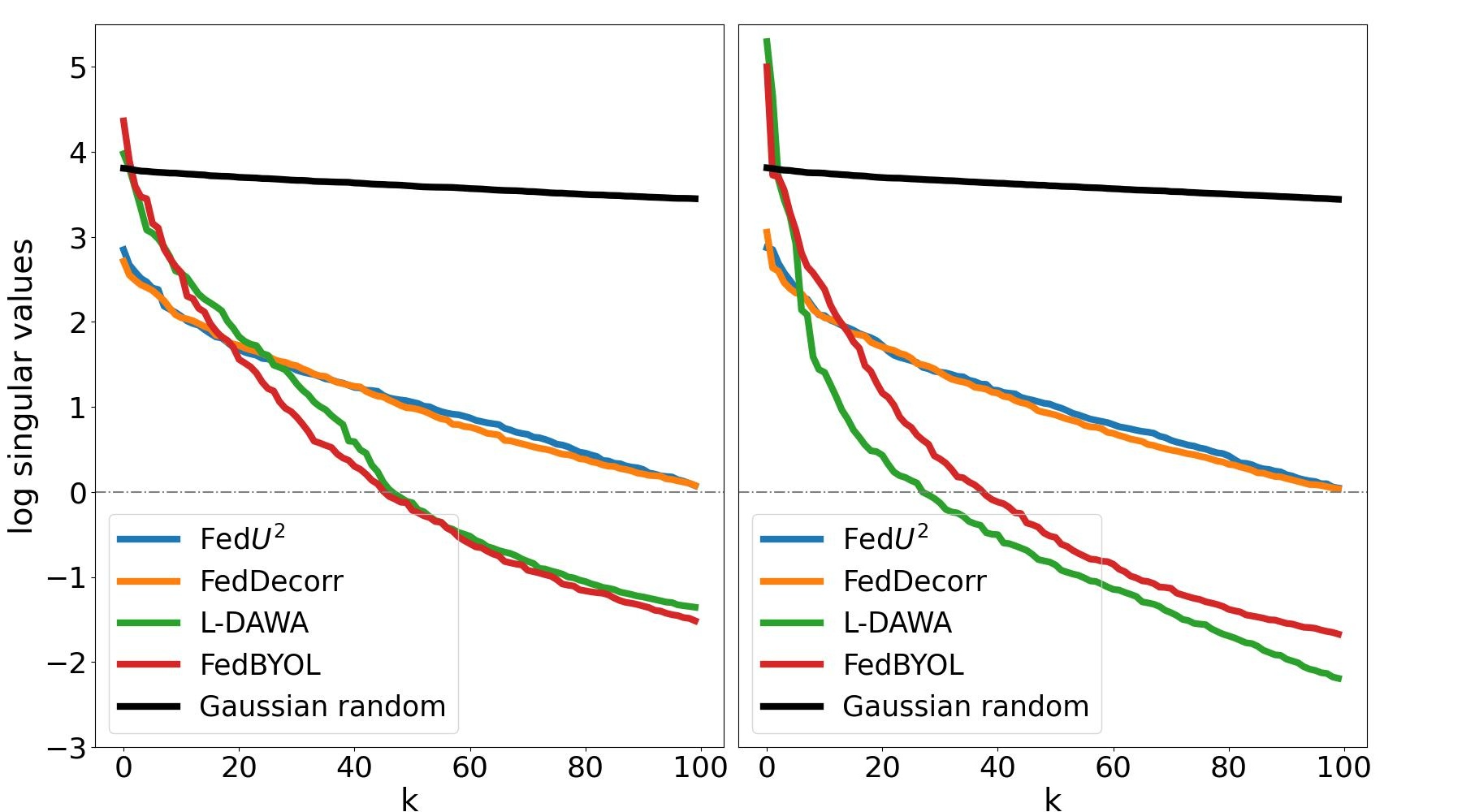}
% \vspace{-0.4cm}
\caption{Top k log singular values of the covariance matrix of global model (left) and local model (right) representations.}
\label{supp_fig:SVD_local}
\end{figure*}

\begin{figure*}[t]
\centering
\subfloat[FedBYOL]
{
    \includegraphics[scale=.24]{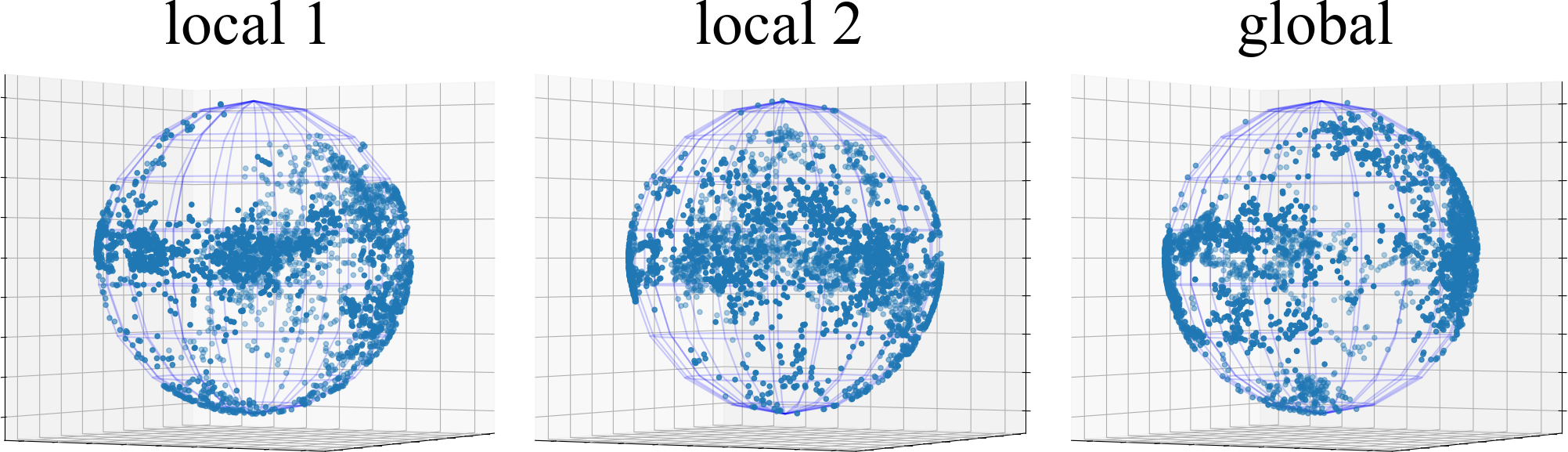}
    \label{supp_fig:collapse_fedbyol}
}\\
\subfloat[FedDecorr]
{
    \includegraphics[scale=.24]{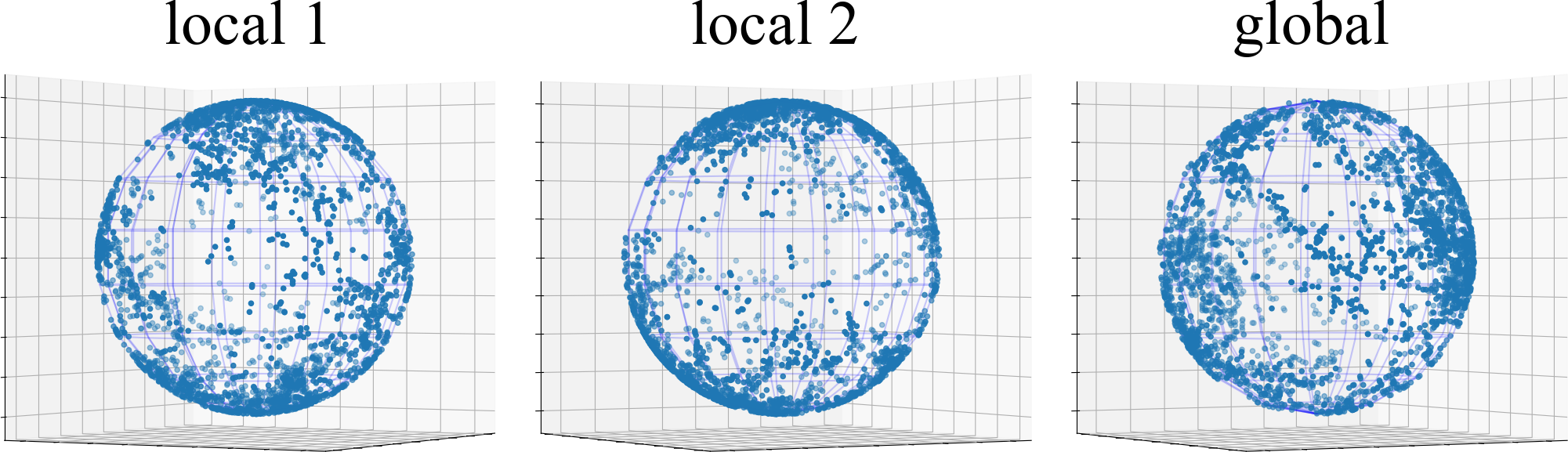}
    \label{supp_fig:collapse_fedDecorr}
}\\
\subfloat[L-DAWA]
{
    \includegraphics[scale=.24]{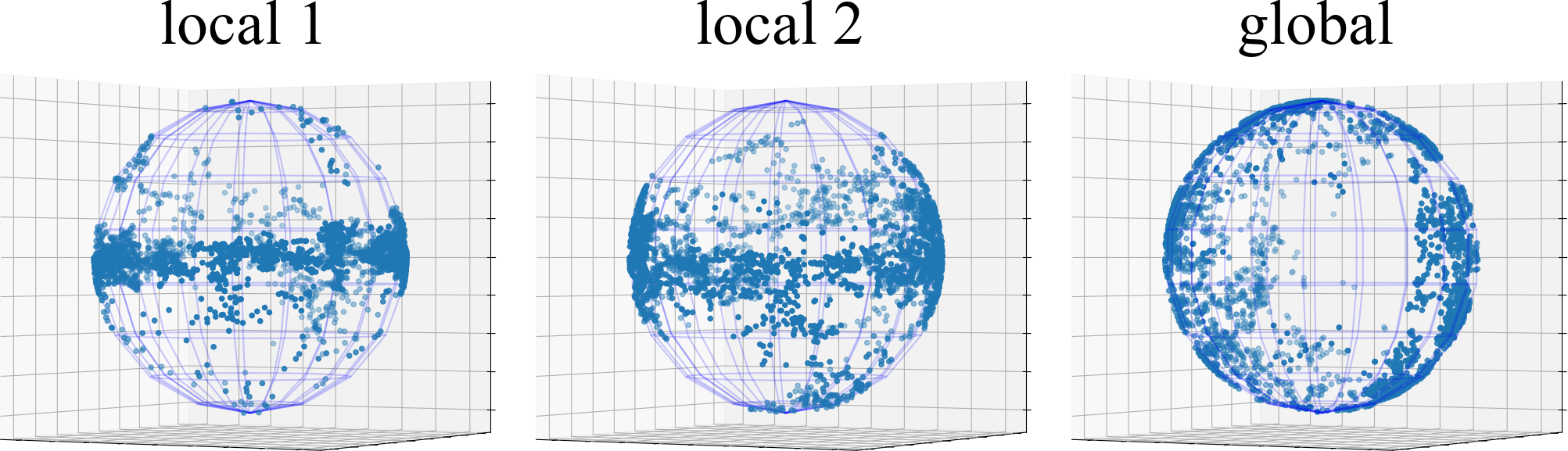}
    \label{supp_fig:collapse_ldawa}
}\\
\subfloat[\modelname]
{
    \includegraphics[scale=.24]{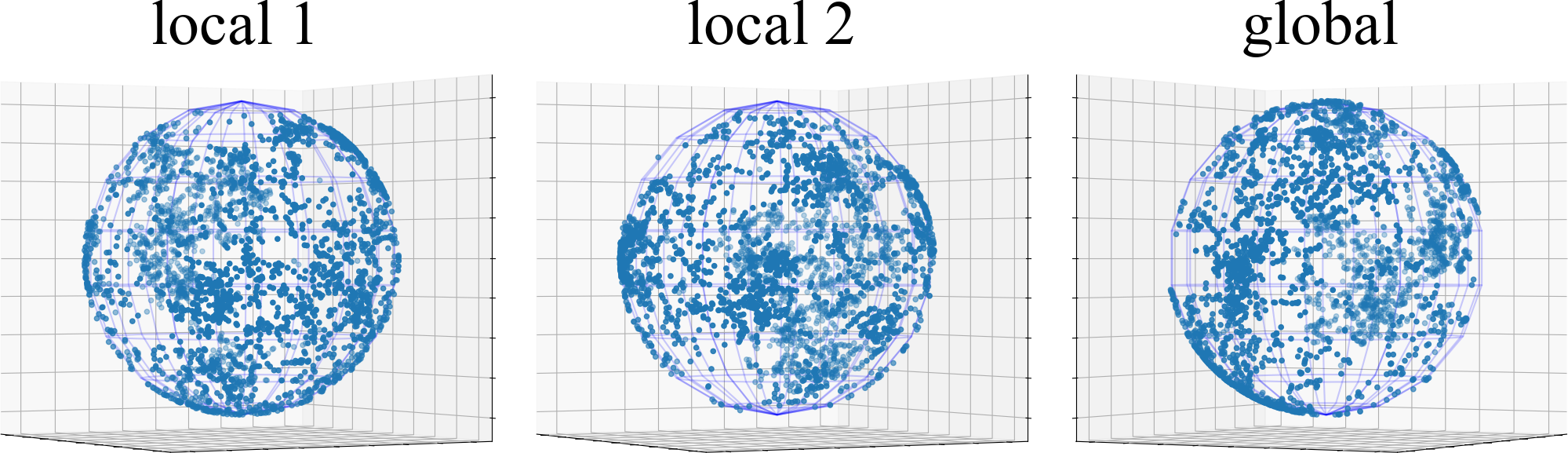}
    \label{supp_fig:collapse_feduu}
}
\caption{The representations collapse issue on the sphere using BYOL model (on CIFAR10 $\alpha=0.1$ Cross-silo). The more blank representation space, the more severe collapse issue is.}
\label{supp_fig:collapse}
\end{figure*}

\begin{figure*}[t]
\centering
\subfloat{
\centering
\includegraphics[width=\linewidth]{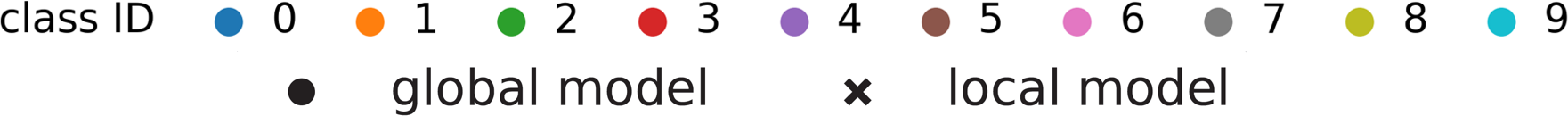} %
}\\
\subfloat[FedBYOL]{
\centering
\includegraphics[width=0.52\linewidth]{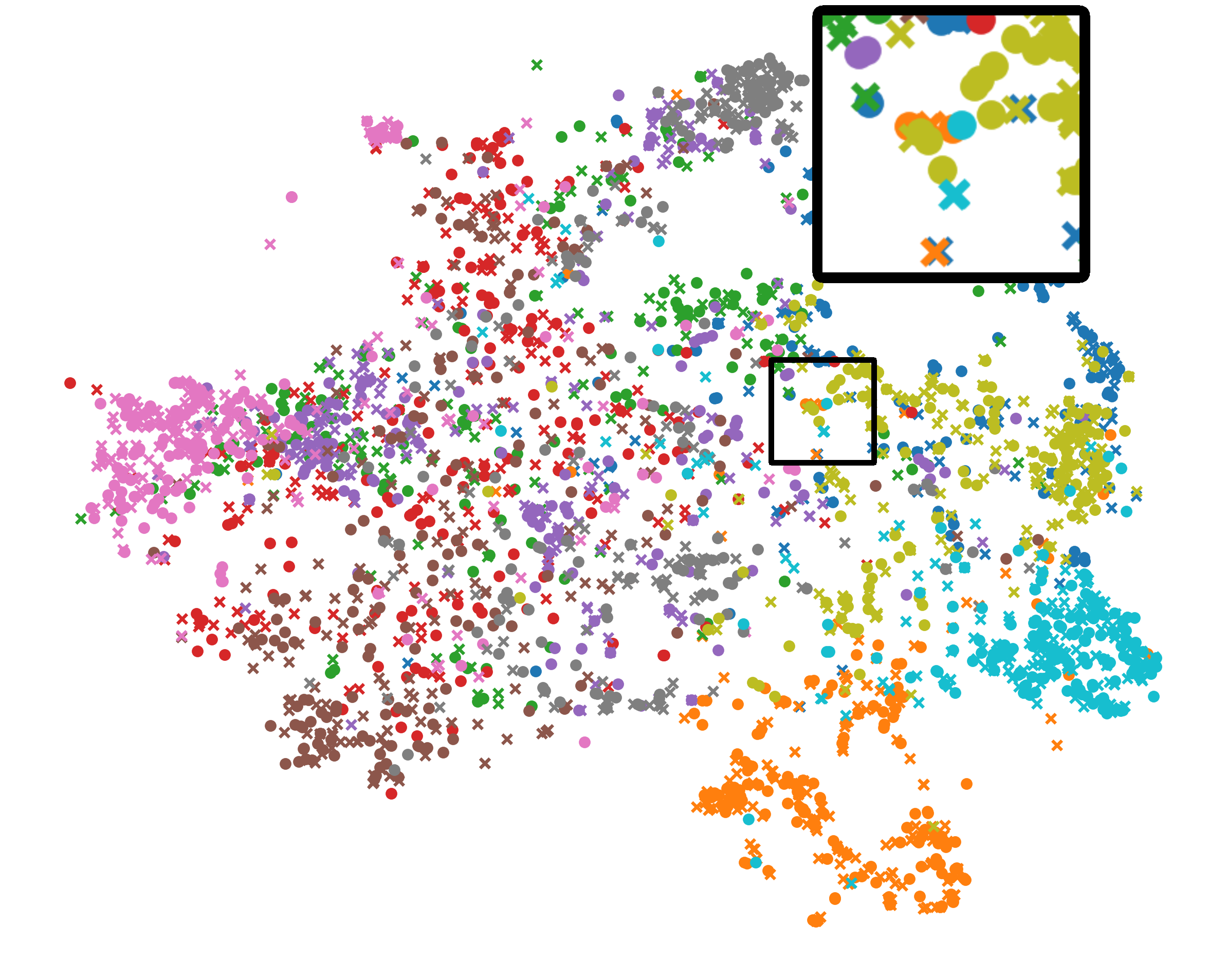} %
}
\subfloat[FedDecorr]{
\centering
\includegraphics[width=0.52\linewidth]{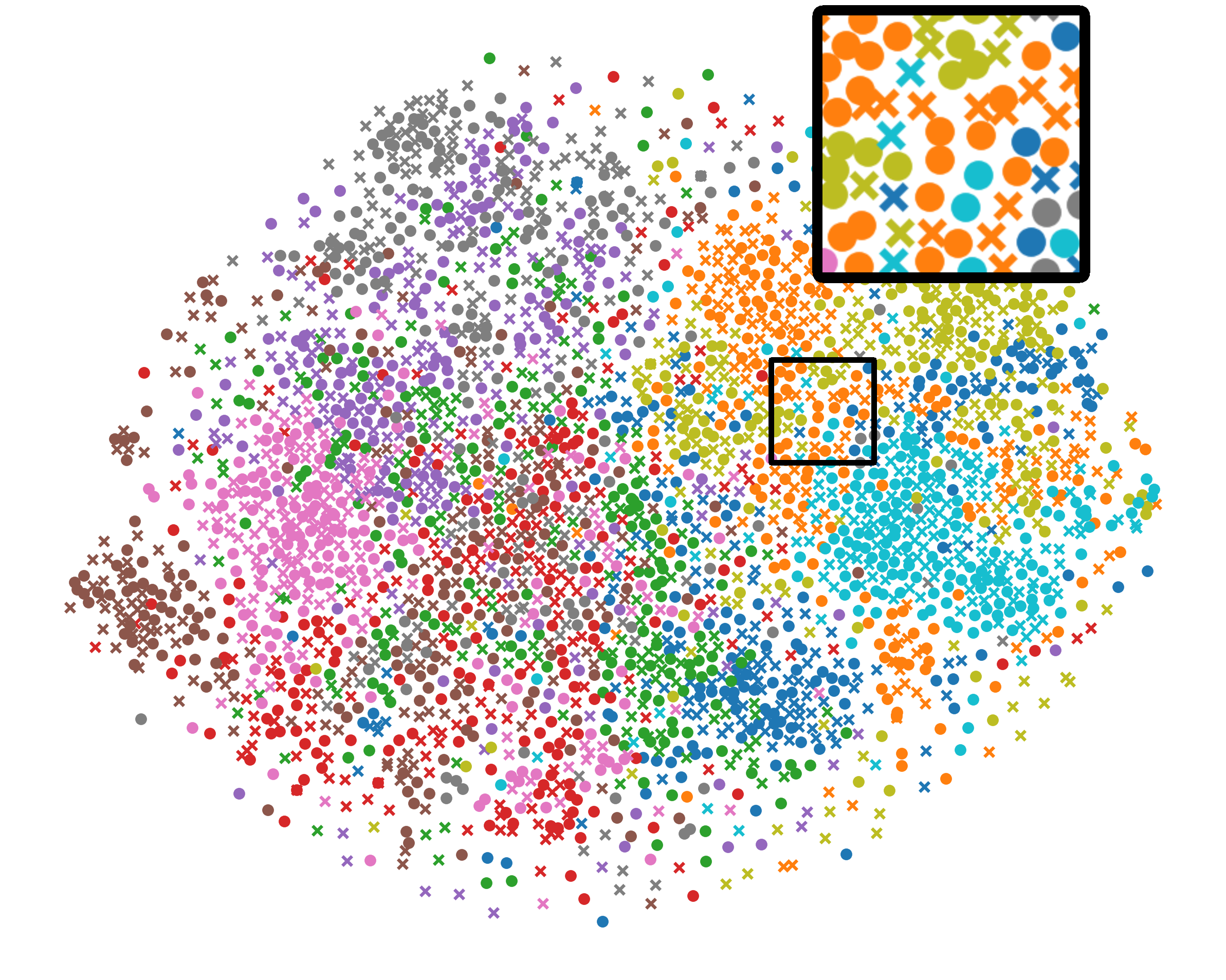} %
}\\
\subfloat[L-DAWA]{
\centering
\includegraphics[width=0.52\linewidth]{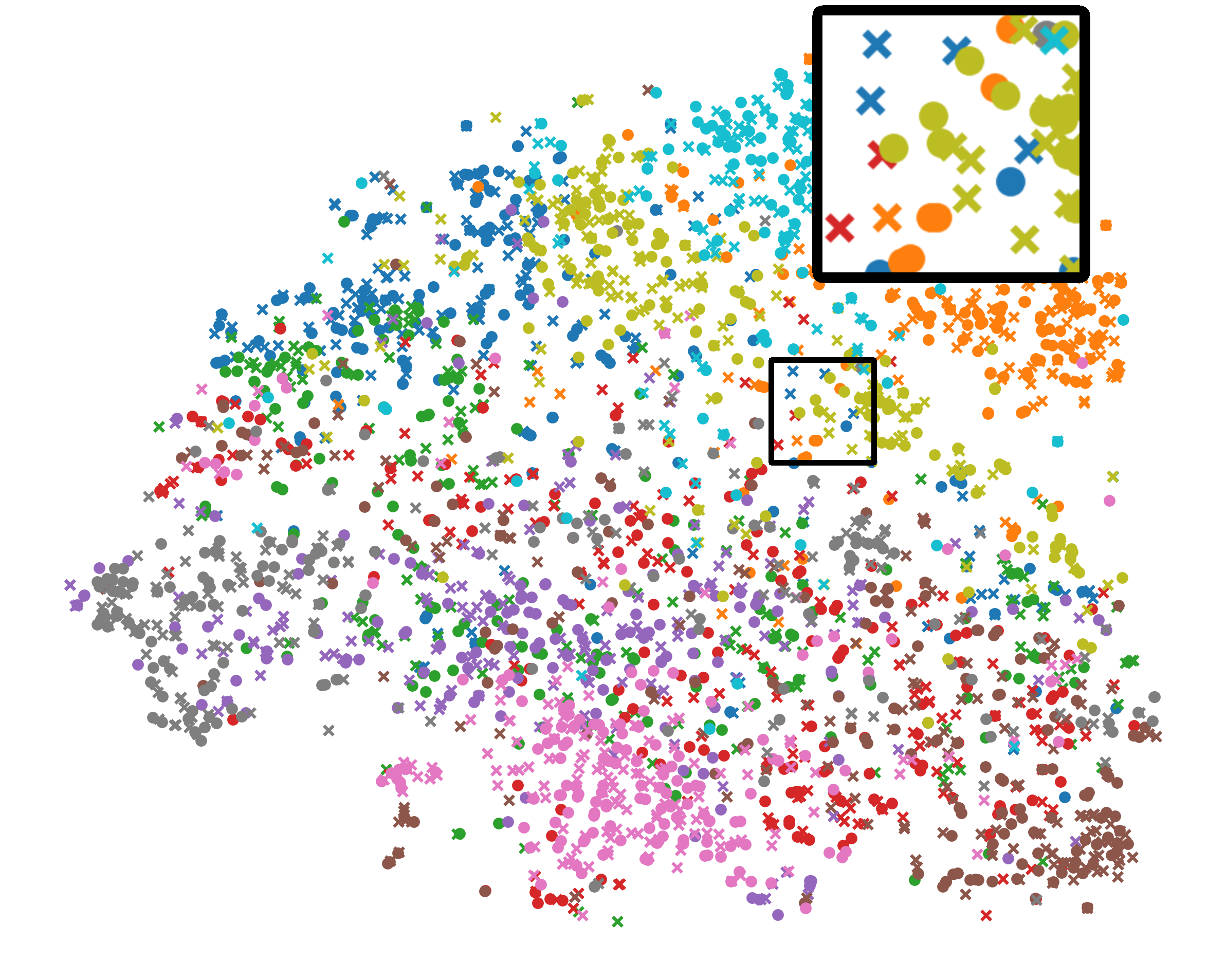} % 
}
\subfloat[\modelname]{
\centering
\includegraphics[width=0.52\linewidth]{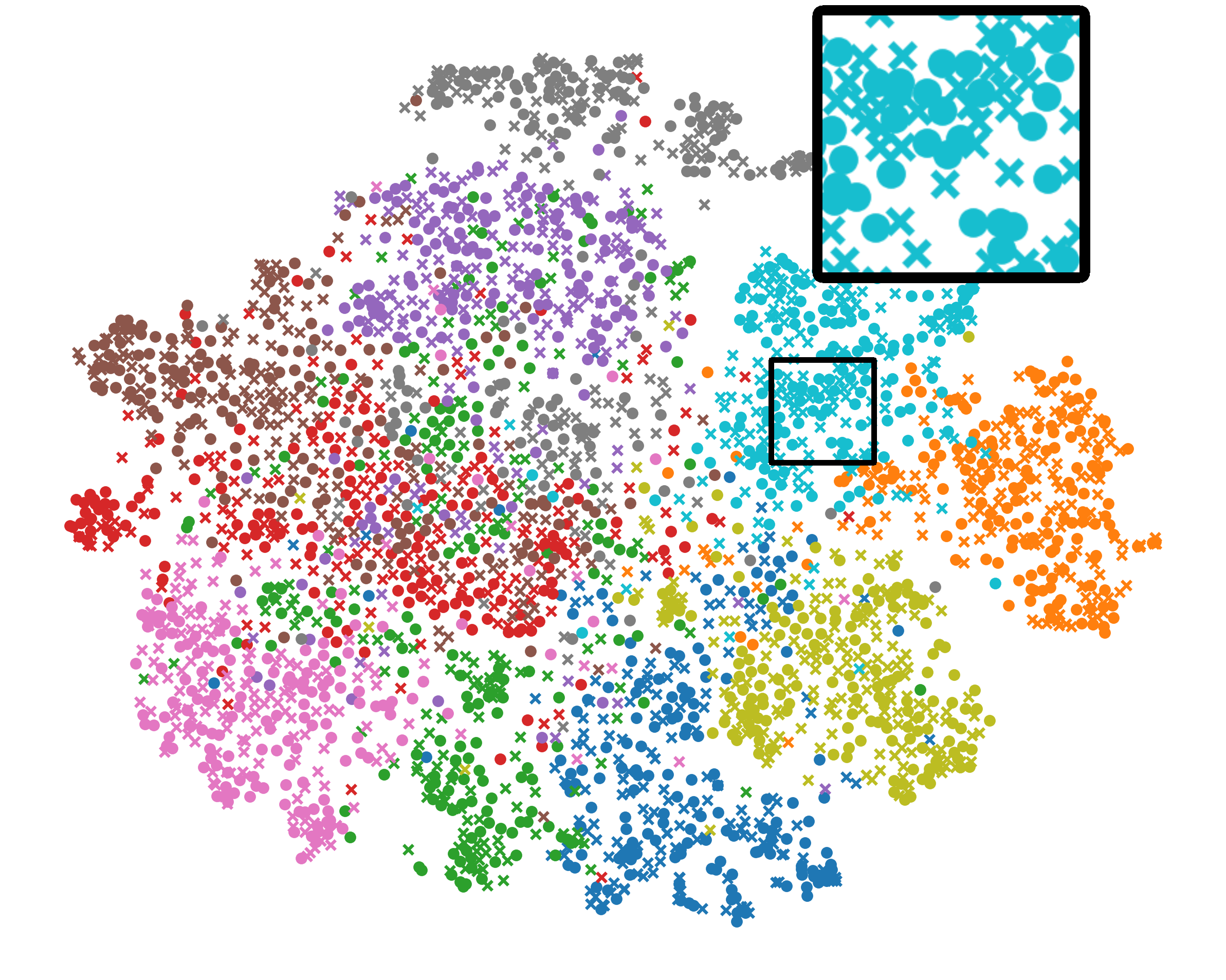} %
}
\caption{The distributions of 
data representations 
using global and local BYOL model (on  CIFAR10 $\alpha=0.1$ Cross-silo).}
\label{supp_fig:unified}
\end{figure*}

In our main paper, we depict the top-k singular values of covariance matrix representations in Fig.~(3), the corresponding 3-D representation in Fig.~(4), and the distribution of data representation in Fig. (5) between global model and randomly sampled local models.
The purpose of the above figures is to illustrate the representation enhancement of \modelname.
In Fig.~\ref{supp_fig:SVD_local}-\ref{supp_fig:unified}, we enlarge these figures to explore the detailed comparisons.
In terms of Fig.~\ref{supp_fig:unified},  \modelname~keeps the unified representation between global and local models as well as clearer decision boundary for each class.

\end{document}